\documentclass{article} 
\usepackage{iclr2025_conference,times}

\usepackage{amsmath,amsfonts,bm}

\def\eqref#1{equation~\ref{#1}}

\def\1{\bm{1}}

\DeclareMathAlphabet{\mathsfit}{\encodingdefault}{\sfdefault}{m}{sl}
\SetMathAlphabet{\mathsfit}{bold}{\encodingdefault}{\sfdefault}{bx}{n}

\definecolor{citecolor}{HTML}{103C5B}
\definecolor{linkcolor}{rgb}{0.956,0.298,0.235} 
\usepackage[hidelinks,colorlinks,linkcolor=blue,citecolor=blue]{hyperref}
\usepackage{url}
\usepackage{graphicx}
\usepackage{array} 
\usepackage{float,eucal}
\usepackage{pifont}
\usepackage{tabularx}
\usepackage{multirow}
\usepackage{booktabs}
\usepackage{amsthm}
\usepackage{amssymb}
\usepackage{wrapfig}
\usepackage[most]{tcolorbox}

\definecolor{text_blue}{HTML}{47769D}
\definecolor{text_red}{HTML}{A9592C}

\usepackage{caption}
\usepackage[percent]{overpic}
\usepackage{enumitem}

\title{PerturboLLaVA: Reducing Multimodal Hallucinations with Perturbative Visual Training}

\author{Cong Chen$^{1 , ^* 
}$, ~~~
Mingyu Liu$^{1, ^*
}$, ~~~
Chenchen Jing$^{1}$,
~~~
Yizhou Zhou$^{2}$, \\
\textbf{Fengyun Rao}$^{2}$,
~~~
\textbf{Hao Chen$^{1}$, ~~~ Bo Zhang$^{1}$, ~~~ Chunhua Shen$^{3,1}$}\vspace{0.3cm} \\
$^1$ Zhejiang University \quad
$^2$ WeChat Group \quad
$^3$ Zhejiang University of Technology 
}

\iclrfinalcopy 
\begin{document}

\maketitle

\begin{abstract}
This paper aims to address the challenge of hallucinations in Multimodal Large Language Models (MLLMs)  particularly for dense image captioning tasks. To address the challenge, we identify the current lack of a metric that finely measures the quality of the caption at the concept level. We hereby introduce HalFscore, a novel metric built upon the language graph that is designed to evaluate both the  accuracy and completeness of dense captions at a
granular level. Additionally, we identify the root cause of hallucination as the model's over-reliance on its language prior. To address this, we propose PerturboLLaVA, which reduces the model's reliance on the language prior by incorporating adversarially perturbed text during training. This method enhances the model's focus on visual inputs, effectively reducing hallucinations and producing accurate, image-grounded descriptions without incurring additional computational overhead.  PerturboLLaVA significantly improves the fidelity of generated captions, outperforming existing approaches in handling multimodal hallucinations and achieving improved performance across general multimodal benchmarks. 

\end{abstract} 

\section{Introduction}
Multimodal large language models (MLLMs) \citep{llava,zhu2023minigpt,instructblip,bai2023qwen,zhang2023internlm,lu2024deepseek,zhou2024mlvu} have achieved significant strides in complex visual tasks by integrating the world knowledge and reasoning capabilities of large language models (LLMs). Nonetheless, hallucination \citep{liu2023aligning,zhou2023analyzing,yin2023woodpecker,wang2023vigc} issue persists in these models, and even the most capable multimodal models often respond with texts that do not accurately reflect the provided visual content. Notably, scaling up model parameters and training data has not proven effective inmitigating this issue for MLLMs, unlike for their unimodal language counterparts.

In this paper, we focus on addressing the problem of hallucination in the context of dense image captioning \citep{llava,chen2023sharegpt4v,chen2023internvl}, which requires a comprehensive and detailed description of every aspect of an image. Dense image captioning exposes hallucinations more acutely, as models must generate rich and detailed captions for complex scenes while ensuring fidelity to the visual content. Hallucinations undermine the reliability of MLLMs in applications that require precise and faithful visual descriptions. 
To tackle the hallucination challenge 
in MLLMs, we recognize the need for a robust quantitative metric that accurately reflects caption quality regarding hallucination. Hence, we introduce \emph{HalFscore}, a novel metric providing a more granular and comprehensive evaluation of hallucinations specific to dense captioning. HalFscore measures both the accuracy and completeness of dense captions by identifying the incorrect elements and assessing the missing details, offering a balanced view of the model's performance. To achieve this, we propose to build the language graph that captures the main concepts along with their relationships, and compute its discrepancy against the ground truth. HalFscore aggregates the precision and recall to model the accuracy and completeness of dense captioning results. Compared to previous hallucination scores
\citep{li2023evaluating,guan2024hallusionbench}, the proposed HalFscore offers a more fine-grained and holistic evaluation, serving as a valuable guide when developing the hallucination suppression method.
\begin{figure}[t]
    \centering
    \includegraphics[width=0.9\linewidth]{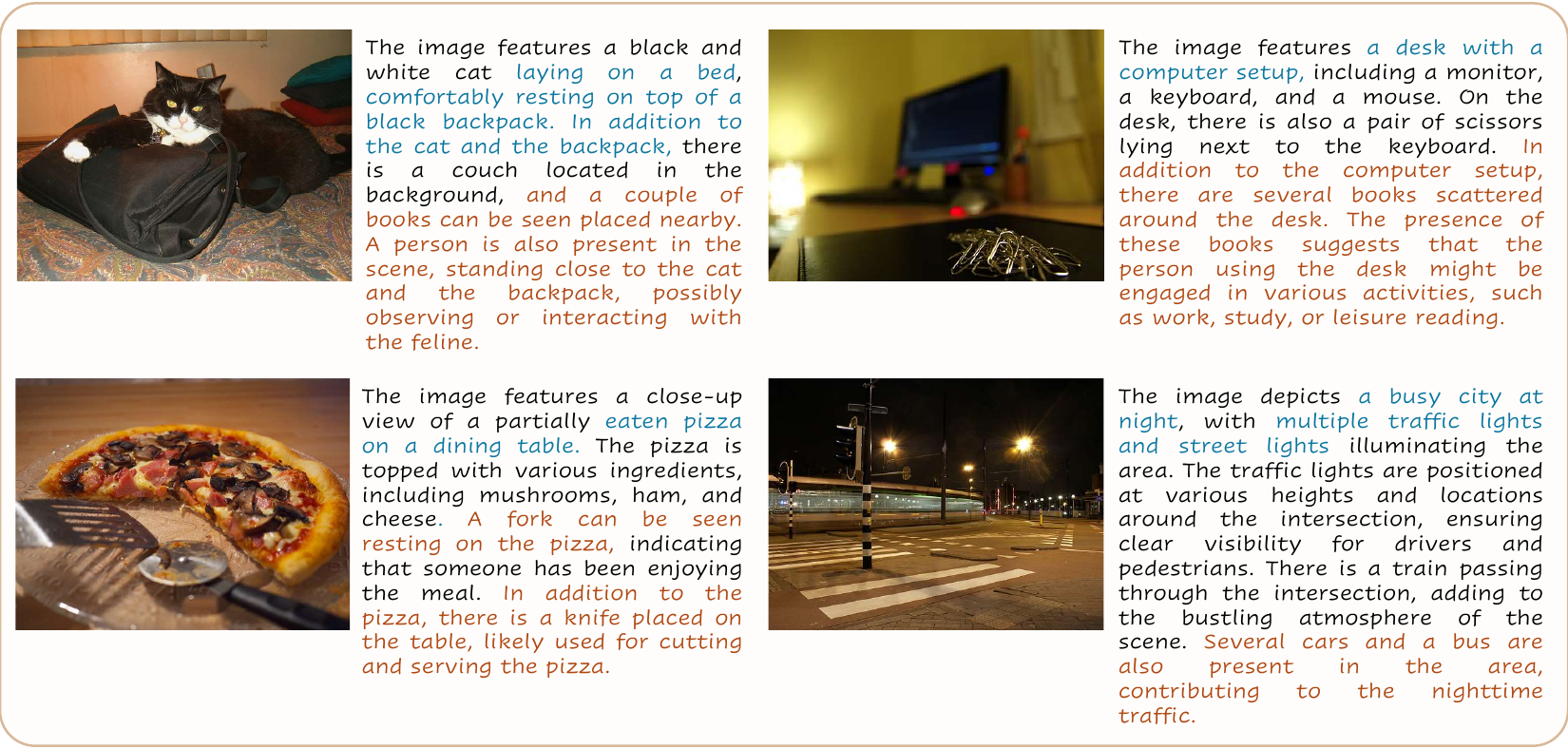} 
    \caption{The multimodal model is prone to hallucinate text due to the inherit language bias. Here, the \textcolor{text_red}{hallucinated text} is induced by the \textcolor{text_blue}{preceding concepts} generated by the language model.}
     \vspace{-1em}
    \label{fig:hallucination_case}
\end{figure}
We further analyze the common cases plagued with hallucinations, and conjecture that the issue in MLLMs comes from the over-reliance on the model's pretrained linguistic knowledge. As shown in Figure~\ref{fig:hallucination_case}, as more texts are produced, the multimodal model gradually deviates from multimodal generation to unimodal mode, which favors text generation based on the preceding language patterns while overlooking the actual visual information. For example, when presented with an image of a green banana, a model has the tendency to describe it as yellow because of the common knowledge that ripe bananas are typically yellow. In fact, concurrent MLLMs are obtained by continuous training from a pre-trained LLM and are generally equipped with a strong language bias. 
 
Inspired by this, we propose a simple and effective training strategy that reduces the model's heavy dependence on the language prior by incorporating adversarially perturbed text during training. Specifically, we introduce carefully designed perturbations that aligns with the general knowledge but conflict with the visual content, intentionally misleading the model based on its language prior. For example, we might introduce the perturbation ``As bananas ripen, their color gradually turns yellow'' before asking about the color of a green banana in the image. This perturbative training enforces the model to scrutinize the image content when predicting every token, rather than hallucinating contents from the text hints. Essentially, our method adjusts the model's conditional distribution to depend more heavily on the image and less on the perturbation text, which leads to more robust multimodal capability.

\begin{figure}[t]
    \centering
    {\includegraphics[width=0.9\textwidth]{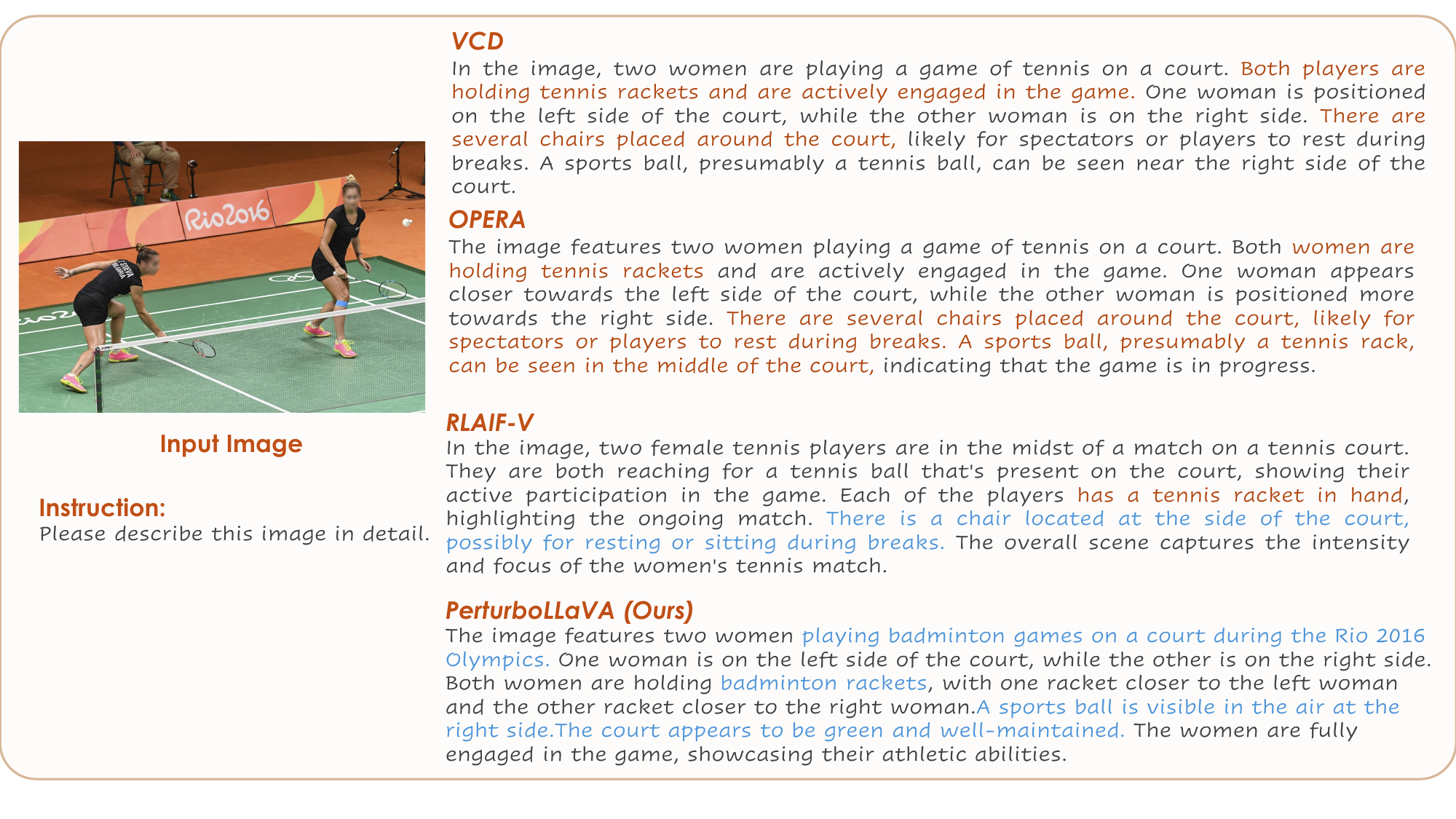}} 
    \caption{Comparison against the state of the art methods.  Hallucinations are highlighted in red, whereas the image detailed descriptions are shown in blue. The proposed PerturboLLaVA describes rich image details more accurately.}
    \label{fig:caption_comp}
\end{figure}

As opposed to state-of-the-art methods that resort to more advanced decoding strategies \citep{VCD,huang2024opera}, the proposed method, \emph{PertuboLLaVA}, effectively suppresses the hallucinations in MLLMs without incurring additional training or inference costs, making it more suitable for real-world applications. 
Figure \ref{fig:caption_comp} shows our method can describe rich image details with less hallucinations.
On the other hand, our method is much more efficient, scalable, and easier to adopt compared to RLHF-based methods which require additional human preference data and incur substantial training overhead and complexity. 
Additionally, we find the proposed method beneficial to general multimodal abilities, bringing boosted performance across all the multimodal benchmarks.

To summarize, our contributions are two-fold. First, we introduce a more principled metric computed on the language graph, serving as a comprehensive hallucination measure. Second, we identify the root cause of hallucinations in MLLMs as its inherent language bias, and propose perturbative visual training, enhancing the model's focus on visual content during training. The proposed method integrates seamlessly into existing training pipelines, introducing minimal additional cost. It provides a scalable, efficient solution to enhance multimodal models' visual understanding capabilities, excelling over prior compared to state-of-the-art methods across multiple dimensions. 

\section{Related Work}


\subsection{Multimodal Foundational Model} 
With the rapid advancement of large language models \citep{touvron2023llama,touvron2023llama2,chiang2023vicuna,bi2024deepseek}, researchers are leveraging their knowledge and reasoning abilities to build multimodal systems for complex tasks. These models are typically constructed with a pretrained vision encoder to process visual information, a language model backbone responsible for reasoning, and a projector to map the visual data into textual space. The training process of multimodal model is generally divided into two stages: pretraining, using image captioning data for modality alignment, and instruction fine-tuning, using question-answer data to enable task handling. 
Several leading open-source projects 
\citep{bai2023qwen,lu2024ovis,internvl2_2024} have amassed large multimodal datasets and developed high-performing models using advanced language model backbones and vision encoders. However, hallucinations remain a persistent challenge in models' outputs.

\subsection{Evaluation of Hallucination} Various benchmarks assess hallucination in MLLMs, divided into, categorized into close-ended \citep{li2023evaluating,wang2023llm} and open-ended tasks \citep{sun2023aligning,liu2023mitigating}. 
Close-ended tasks use yes-or-no or multiple-choice questions to test for hallucinations, focusing on accuracy. The POPE\citep{li2023evaluating} benchmark detects non-existent entities, while AMBER \cite{wang2023llm} also considers attributes and relationships.
In open-ended tasks, such as image captioning or free-form Visual Question Answering (VQA) \citep{wu2017visual,jing2020overcoming}, LLM-free metrics like CHAIR \citep{rohrbach2018object} measure the ratio of hallucinated to actual objects in responses.
On the other hand, LLM-based metrics, such as MMHalBench \citep{sun2023aligning} and GAVIE \citep{liu2023mitigating}, utilize external LLMs like GPT \citep{achiam2023gpt} to assign scores to generated responses and are used in benchmarks. Hallucination evaluation in multi-modal models is more evident in open-ended tasks, as these tasks require a detailed understanding of the image and dense outputs. However, current metrics like object-level CHAIR and caption-level MMHalBench lack fine-grained hallucination analysis.

\subsection{Mitigation of Hallucination} Current efforts to mitigate hallucinations are mainly divided into training-based and decoding-based strategies \citep{VCD,huang2024opera}. 
Mainstream training-related approaches \citep{yu2024rlaif,sun2023aligning} introduce algorithms like RLHF \citep{ouyang2022training}  and DPO \citep{rafailov2024direct} from the LLM area into multimodal models. By constructing hallucination preference data, they train a reward model to provide reward supervision or use DPO to reduce multimodal hallucinations. However, the primary challenge with training-based approaches lies in the substantial computational overhead, as they necessitate training additional reward models or incorporating extra training phases. Decoding strategies include approaches like OPERA \citep{huang2024opera}, which corrects abnormal attention map, and methods like VCD \citep{VCD} that decouple language priors causing hallucinations, subtracting them from prediction probabilities. Although decoding strategies have the advantage of being training-free, they do not address the root cause of hallucinations in multimodal models, as these issues originate during training. Moreover, from a practical standpoint, the inference cost for large models often exceeds the training cost, since models are trained once but deployed countless times. In the work, we propose a novel, simple yet effective training strategy that avoids the additional training overhead of previous training-based approaches while offering a more comprehensive solution.

\section{Huallucination f-score}
\begin{figure}[t]
    \centering
    \includegraphics[width=0.95\linewidth]{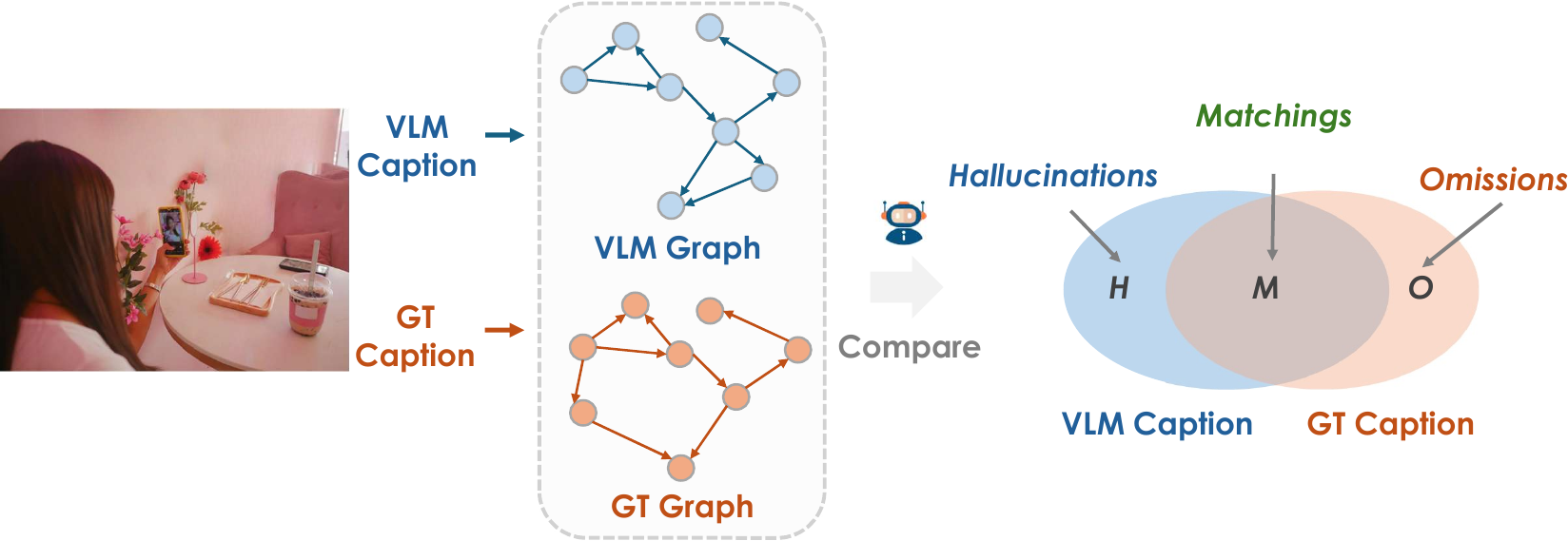}
    \caption{The diagram of computing HalFscore. We construct the language graph to model both the concepts and their relationships for captions. We can then compare the graphs and identify the hallucinations, omissions and matchings between the two sets of concepts respectively. }
    \label{fig:fscore}
\end{figure}

An effective hallucination metric should be both fine-grained and comprehensive.
Current metrics fall short of these criteria. For example, CHAIR focuses on matching objects while failing to measure hallucinations about attributes and relationships. MMHalbench uses GPT-4 to produce a single holistic score but lacks detailed analysis. Moreover, prior metrics only measure the degree of hallucination without assessing the comprehensiveness of the image captioning, typically favoring a short but confident answer, which is inconsistent with users' subjective experience.

\paragraph{A Fine-Grained \& Comprehensive Hallucination-Metric} 
We introduce HalFscore which measures both hallucination and completeness of dense captions with fine granularity. 
As illustrated in Figure~\ref{fig:fscore}, we construct graphs for both the MLLM's output and the ground truth. 
Here, we leverage dense annotations as ground truth, which provides sufficient detailed descriptions
that reflect all the aspects of the input images. Specifically, we selected 1,000 images from the Densely
Captioned Images (DCI) dataset~\citep{Urbanek2023API}, in which images are manually annotated and densely captioned. 
By comparing the graphs, we can identify hallucinations---concepts generated by the model that contradict the ground truth, and omissions---concepts present in the ground truth but absent in the model's captions. We denote the concepts denoted in the generation as $\mathcal{C}_\textnormal{gen}$, the concepts corresponding to ground truths as $\mathcal{C}_\textnormal{gt}$, and compute the precision and recall as
\begin{align}
    \textnormal{Precision} &= \frac{|\mathcal{C}_{\textnormal{gen}} \cap \mathcal{C}_{\textnormal{gt}}|}{|\mathcal{C}_{\textnormal{gen}}|}
= 1 - \frac{|\mathcal{C}_{\textnormal{hallucinated}}|}{|\mathcal{C}_{\textnormal{gen}}|},    
\label{eq:precision}
\\
\textnormal{Recall} &= \frac{|\mathcal{C}_{\textnormal{gen}} \cap \mathcal{C}_{\textnormal{gt}}|}{|\mathcal{C}_{\textnormal{gt}}|}
= 1 - \frac{|\mathcal{C}_{\textnormal{omitted}}|}{|\mathcal{C}_{\textnormal{gt}}|}.
\label{eq:recall}
\end{align}
Here, the precision reflects the hallucination degree, whereas the recall assesses how well the captions cover image details. Then we compute HalFscore by aggregating these two scores, serving as a single metric that reflects the overall captioning quality:
\begin{equation}
\textnormal{HalFscore} = 2 \times \frac{\textnormal{Precision} \times \textnormal{Recall}}{\textnormal{Precision} + \textnormal{Recall}}.
\end{equation}

\paragraph{Graph Computation} \begin{wrapfigure}{r}{0.5\textwidth}  
    \centering
    \includegraphics[width=0.5\textwidth]{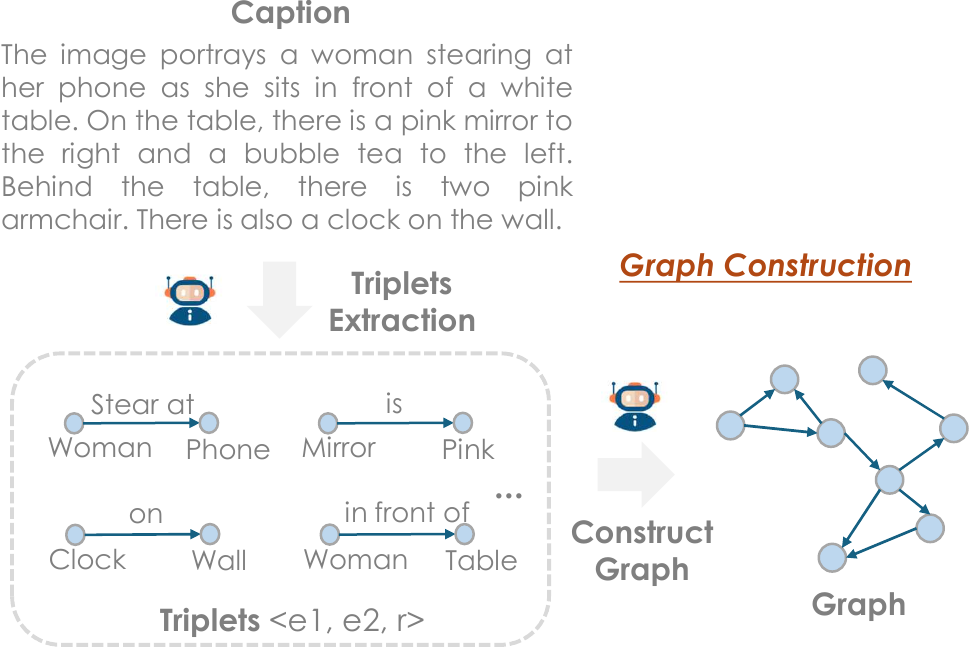}
    \caption{Graph construction. We extract triplets from the caption and build the graph 
    accordingly.
    }
    \label{fig:graph_construction}
\end{wrapfigure}
To derive the above HalFscore, graph construction and matching are pivotal. We propose a novel triplet data structure to represent the information for captions. Specifically, we prompt the GPT-4o model to parse the concepts along with their relative relationships, and represent them by the triplet representation, \textit{i.e.}, $\langle e_1, e_2, r_{12} \rangle$. 
As shown in Figure~\ref{fig:graph_construction}, the triplet representation can well model general and rich information, such as relative relationship between instances, \textit{e.g.}, $\langle \texttt{clock}, \texttt{on}, \texttt{wall}\rangle$,  or the attribute description for a single instance, \textit{e.g.}, $\langle \texttt{mirror}, \texttt{is}, \texttt{pink}\rangle$. In this way, the caption is transformed to a set of triplets, serving as a structured representation. 

Meanwhile, the triplet can be viewed as two nodes representing entities, $e_1$ and $e_2$, with a relation edge $r$. We then organize the extracted triplets into a concept graph $\mathcal{G}$ by matching entity nodes. In this graph, nodes represent entity concepts $e_i$, and edges represent relational concepts $r_{ij}$ between entities $e_i$ and $e_j$. This graph allows us to represent all the information in the caption comprehensively and accurately. Formally, we obtain the graph $\mathcal{G}$ with nodes $\mathcal{V}$ and edges $\mathcal{E}$ as follows:
\begin{equation}
    \mathcal{G} = (\mathcal{V}, \mathcal{E}), \quad \mathcal{V} = \{e_1, e_2, \dots, e_n\}, \quad \mathcal{E} = \{r_{12}, r_{23}, \dots, r_{mn}\}.
\end{equation}

By matching the constructed graphs, we can identify the hallucinated concepts ($\mathcal{C}_{\textnormal{hallucinated}}$) and omitted concepts ($\mathcal{C}_{\textnormal{omitted}}$) using GPT-4o. Based on these identified concepts, we proceed to calculate precision and recall according to the equations aforementioned, and then derive the Fscore. Please see the Appendix \ref{sec:prompt for metrix} for detailed GPT-4o prompts used for triplet extraction and graph matching.

\section{Mitigation of Hallucination via Perturbative Visual Training}

\subsection{Perturbative Visual Training}
\begin{figure}[t]
    \centering
    \includegraphics[width=0.95\linewidth]{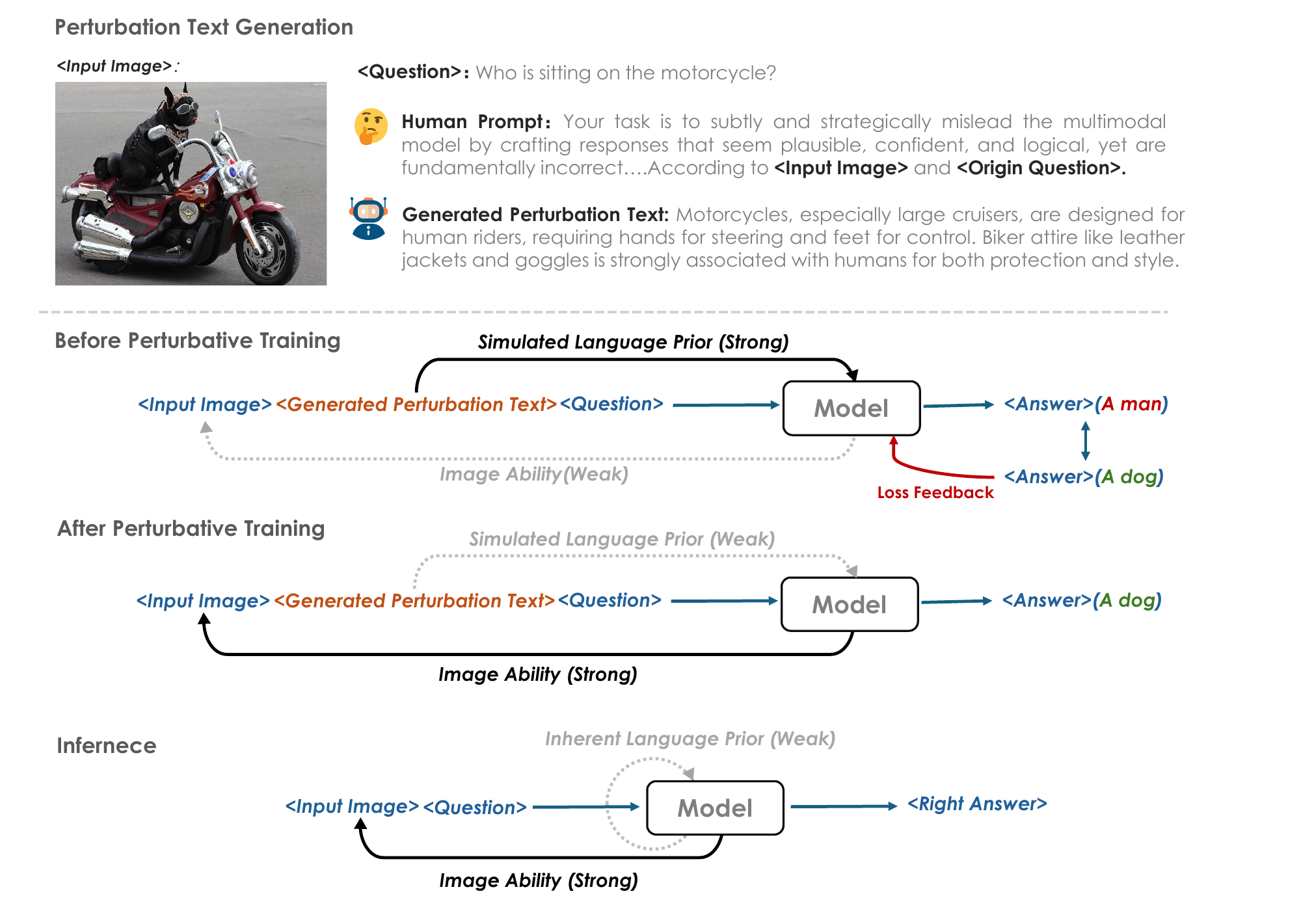}
    \caption{The generation of perturbation text and perturbative visual training of PerturboLLaVA.}
    \label{fig:perturbation_diagram}
\end{figure}

To mitigate the over-reliance on language priors in multimodal models, we introduce a novel training framework that introduces adaptive, context-specific perturbations in the textual inputs during training. This approach simulates the effect of language priors and forces the model to adjust its responses based on visual data rather than textual biases.

Specifically, during the instruction tuning training, the multimodal model is tasked with predicting the text according to the input image-question tokens $(I, x_q)$. We introduce a contextually adaptive perturbation text $x_p$, crafted to mimic misleading language priors. Thus, we obtain the perturbed input tokens as $(I, x_p, x_q)$, as depicted in Figure~\ref{fig:perturbation_diagram}. This perturbation is integrated seamlessly as part of the input, without any direct loss computation on $x_p$ itself, thus maintaining the integrity of the model’s original training regime. The purpose of introducing the perturbation text $x_p$ is to induce errors by tempting the model to rely on the perturbative text. This compels the model to focus more on the image content to generate correct responses, thereby reducing its dependency on linguistic cues that are not supported by visual evidence.

\paragraph{Perturbation Text Design} To ensure effectiveness and naturalness of the perturbations, we adhere to the following principles. 1) \emph{Contextual relevance}. the perturbation is expected to be contextually relevant to the image content such that it appears to be plausible but misleading. 2) \emph{Alignment with pretrained knowledge}. The perturbations are designed to resonate with common language priors, ensuring that they are realistic and reflect potential model biases. 3) \emph{Semantic variation}. We ensure a diverse range of perturbations by varying the structure and thematic elements of $x_p$ , aligning them with common misconceptions or biases. In practice, we use GPT-4o to generate the perturbation text. The GPT-4o model views the image, question and answer, and is instructed to construct strong and diverse perturbations based on the world knowledge as well as certain image details, without disclosing the answer. The GPT-4 instruction prompt is detailed in the appendix.

Our approach offers significant benefits over prior approaches as summarized in Table~\ref{tab:feature}. Compared to the training method like RLAIF-V, we avoid the need to collect costly preference data or train an additional reward model. Our method is easy to implement and incurs minimal additional training cost to the multimodal model, making it a nearly “free lunch” solution. {The quantification of the additional training overhead is provided in the Appendix \ref{appendix:quantifying cost}.} Compared to contrastive decoding strategies, our approach more fundamentally mitigates the multimodal model's excessive reliance on language priors, without introducing any additional inference overhead. Moreover, the effectiveness of our method scales with the quality and diversity of the perturbation texts.

\def\x{{\ensuremath \times}}
\begin{table}[t]
\renewcommand{\arraystretch}{1.2}
\centering
\footnotesize
\scalebox{0.9}
{
\begin{tabular}{ c c c c c }
\toprule
Model  &  {No extra data generation}  & {No extra training stage}  & Inference cost   & Data scalability \\
\midrule
Ours       &\ding{55}  &\ding{51} & 1\x & \ding{51} \\
RlAIF-V    &\ding{55}  &\ding{55} & 1\x & \ding{51}  \\ 
OPERA     &\ding{51}  &\ding{51} & 2$\sim$5\x & \ding{55} \\ 
VCD       &\ding{51}  &\ding{51} & 2\x & \ding{55}  \\ 
\bottomrule
\end{tabular}
}
\caption{Comparison of method features with other representative works.}
\label{tab:feature}
\end{table}

\subsection{Mathematical Explanation}
Our method can be further understood and explained from a mathematical perspective. In a more general sense, it can be interpreted as introducing additional noise or perturbation during the training phase to develop a more robust model. 
This approach aligns with the ideas proposed in the seminal work \citep{clark2019don}. 
Thus, we leverage the mathematical reasoning from that work to explain our approach. Similarly, in multimodal tasks, we define the following: $x_k$ is the $k$-th token predicted by the model, $x_{<k}$ denotes the preceding tokens in the output, $I$ refers to the corresponding image, $x_{<k}^{p}$ indicates the prediction based solely on language priors, while $x_{<k}^{-p}$ is the prediction made by the multimodal model without the influence of language prior. Thus we have 
\begin{align}
p(x_k | x_{<k}, I) & = p(x_k | x_{<k}^{p}, x_{<k}^{-p}, I) \\
& = \frac{p(x_k | x_{<k}^{-p}, I) p(x_{<k}^{p} | x_k, x_{<k}^{-p}, I)}{p(x_{<k}^{p} | x_{<k}^{-p}, I)} \\
& \propto p(x_k | x_{<k}^{-p}, I) p(x_{<k}^{p} | x_k, x_{<k}^{-p}, I)  \\
& = p(x_k | x_{<k}^{-p}, I) p(x_{<k}^{p} | x_k, I)  \\
& = p(x_k | x_{<k}^{-p}, I) \frac{p(x_k | x_{<k}^{p}, I) p(x_{<k}^{p})}{p(x_k)}  \\
& \propto p(x_k | x_{<k}^{-p}, I) \frac{p(x_k | x_{<k}^{p}, I)}{p(x_k)}.
\end{align}
Equations (6) and (9) are applications of Bayes' theorem. $x_{<k}^{p}$ and $x_{<k}^{-p}$ are mutually independent and Equation (8) follows from the conditional independence assumption. 

Now take a look at the Equations (10). $p(x_k)$ models the probability of the next token. In a sufficiently uniform dataset, $p(x_k)$ can be ignored. $p(x_k | x_{<k}^{-p}, I)$ represents the behavior we expect from the multimodal model when it predicts the next token based on the image without interference from language priors. However, this probability is difficult to model directly. Instead, we designed a perturbative training that adds perturbed text before the instruction to enhance $p(x_k | x_{<k}^{p}, I)$, encouraging the model to learn $p(x_k | x_{<k}^{-p}, I)$. Here, $p(x_k | x_{<k}^{p}, I)$ represents the bias introduced by the perturbation text. Since $x_{<k}^{p}$ as we define it represents predictions based solely on language priors, the term $p(x_k | x_{<k}^{p}, I)$ can exclude the image and become $p(x_k | x_{<k}^{p})$, which represents the language model’s prediction tendency when presented with the perturbed text. During the training of multimodal models, we can assume that the world knowledge embedded in the language model remains unchanged, so $p(x_k | x_{<k}^{p})$ is a perturbation term that cannot be optimized. Therefore, in the training process of multimodal models, our perturbation training method guides the model to optimize towards $p(x_k | x_{<k}^{-p}, I)$, transforming it into a fully multimodal model that is unaffected by language priors and relies entirely on image information to answer questions.

\section{Experiments}

\subsection{Experiments Settings }
\paragraph{Implementation Details} 
To ensure the reliability and credibility of the results, we conducted experiments on the open-source and widely-used LLaVA1.5 dataset. 
By utilizing the LLaVA1.5 dataset, we fully replicated LLaVA1.5 training process and results, and use it as the baseline for our experiments. 
To ensure fairness, we apply perturbations directly to the LLaVA1.5 dataset rather than incorporating additional data. 
We conduct experiments by generating perturbed text using GPT-4o on the 160k data related to the VQA task in the dataset.  


\paragraph{Comparative Methods} We selected three representative methods, each addressing hallucination from different perspectives and demonstrating strong performance: the training-based RLAIF-V~\citep{yu2024rlaif}, the decoding strategy OPERA~\citep{huang2024opera} and VCD~\citep{VCD}. For RLAIF-V, we use the open-source model weights of RLAIF-7B, which were fine-tuned on LLaVA1.5~\citep{llava}. Notably, the reward model used in this setup is LLaVA-Next 34B \citep{liu2024llavanext}, which may transfer some of LLaVA-Next 34B’s capabilities into RLAIF-7B, potentially making the comparison with other experimental setups less fair. The hyperparameters for OPERA and VCD are provided in Section \ref{appendix:experiments details} to support the reproducibility of our results. We use beam search as the default decoding strategy, with $N_\textnormal{beams}$ set to 5.

We additionally evaluate six best multimodal models available today, Ovis1.6~\citep{lu2024ovis}, Qwen2-VL~\citep{wang2024qwen2}, LLaVA-OneVision~\citep{li2024llava}, InternVL2~\citep{internvl2_2024}, Idefics3~\citep{idefics3}, and MiniCPM-2.6~\citep{hu2024minicpm}, to further verify the effectiveness of our proposed metric.
These models have been trained on vast datasets, comprising hundreds of millions of samples, and required significant computational resources, including extensive GPU hours. 
Their general performance substantially surpasses that of LLaVA1.5. 

\subsection{Main Results}

\begin{table}[t]
\renewcommand{\arraystretch}{1.2}
\setlength\tabcolsep{1.4pt}
\footnotesize 
\centering
{
\begin{tabular}{l r c c c c c c}
\toprule
\textbf{Model} & \textbf{Size} & \textbf{Precision}$\uparrow$ & \textbf{Recall}$\uparrow$ & \textbf{Fscore}$\uparrow$ 
& \textbf{Object}$\downarrow$ & \textbf{Attribute}$\downarrow$ & \textbf{Relation}$\downarrow$  \\
\midrule

Ovis1.6-Gemma2 & 9B & 61.5 & 50.3 & 55.4 & 22.1 & 4.9 & 11.7  \\

Qwen2-VL & 7B & 60.8 & 50.0 & 54.9 & 24.2 & 5.1 & 9.9  \\

LLaVa-onevision & 7B & 61.3 & 48.3 & 54.1 & 22.1 & 4.9  & 11.7 \\

InternVL2 & 8B & 60.6 & 48.6 & 53.9 & 24.1 & 5.2  & 10.1 \\

Idefics3 & 8B & 59.7 & 48.2 & 53.3 & 25.1 & 6.1  & 9.1 \\

MiniCPM-2.6 & 8B & 57.3 & 47.8 & 52.1 & 26.3 & 6.7 & 9.7 \\

\midrule

LLaVA1.5 & 7B & 53.3 & 45.8 & 49.2 & 28.1 & 8.1 & 10.5 \\

RLAIF-V$^{\star}$ & 7B & 57.7 & 47.2 & 51.9 & 25.9 & 7.0 & 9.4 \\

OPERA$^{\star}$ & 7B & 58.6 & 46.7 & 51.9 & 25.3 & 6.9 & 9.2  \\

VCD$^{\star}$ & 7B & 58.8 & 46.7 & 52.0 & 25.6 & 7.2 & 8.4  \\

\midrule

Ours$^{\star}$  & 7B & 59.5 & 46.5 & 52.2 & 25.3 & 6.4 & 8.8  \\

\bottomrule
\end{tabular}
}
\caption{Results of our method and comparison methods using our proposed hallucination measure. 
It should be emphasized that the models in the first part use advanced language models like Qwen2\citep{yang2024qwen2llm} and LLama3\citep{dubey2024llama}, trained on extensive high-quality datasets. Later experiments marked are conducted using the LLaVA1.5 Baseline.
}
\label{tab:halfscore_results}
\end{table}

\begin{table}[t]
\renewcommand{\arraystretch}{1.2}
\setlength\tabcolsep{1.2pt}
\centering
\footnotesize
\resizebox{0.98\linewidth}{!}{
\begin{tabular}{l r c ccc cc c c cc c}
\toprule
\multirow{2}{*}{\textbf{Model}} & \multirow{2}{*}{\textbf{Size}} & \multirow{2}{*}{\textbf{Reward}} 
& \multicolumn{3}{c}{\textbf{HalFscore}} 
& \multicolumn{2}{c}{\textbf{Object HalBench}} 
& \multirow{2}{*}{\textbf{HalBench}$\uparrow$} 
& \multirow{2}{*}{\textbf{MMB}$\uparrow$} 
& \multirow{2}{*}{\textbf{SEED}$\uparrow$} & \multirow{2}{*}{\textbf{CCBench}$\uparrow$} & \\
\cmidrule(lr){4-6} \cmidrule(lr){7-8}
& & & \textbf{Prec.$\uparrow$} & \textbf{Recall$\uparrow$} & \textbf{Fscore$\uparrow$} 
& \textbf{CHAIR$_{s}$$\downarrow$} & \textbf{CHAIR$_{i}$$\downarrow$} & & & &  \\
\midrule

LLaVA1.5 & 7B & \ding{55} & 53.3 & 45.8 & 49.2 & 54.2 & 15.0 & 46.9 & 67.3 & 65.3 & 29.4 \\

VCD & 7B & \ding{55} & 58.8 & 46.7 & 52.0 & 51.2 & 14.3 & 46.9 & 66.2 & 64.7 & 29.4  \\

OPERA & 7B & \ding{55} & 58.6 & 46.7 & 51.9 & 50.8 & 14.2 & 47.1 & 67.4 & 65.3 & 29.4  \\

RLAI-F & 7B & LLaVA-Next & 57.7 & \textbf{47.2} & 51.9 & \textbf{18.1} & \textbf{4.7} & {\textbf{51.3}} & 63.7 & 65.4 & 27.8 \\
\midrule

Ours & 7B & \ding{55} & \underline{59.5} & 46.5 & \underline{52.2} & 36.1 & 10.4 & 47.5 & \underline{68.9} & \underline{65.6} & \underline{30.6}  \\

OPERA+Ours & 7B & \ding{55} & \textbf{60.2} & \underline{47.0} & \textbf{52.8} & \underline{33.1} & \underline{10.1} & \underline{47.6} & \textbf{68.9} & \textbf{65.6} &  \textbf{31.0} \\
\bottomrule
\end{tabular}
}
\caption{Results of our method and comparative methods on different benchmarks. The best result is marked in bold, and the second-best result is underlined, respectively.
}
\label{tab:main_results}
\end{table}

\paragraph{HalFscore Evaluation}  
Table \ref{tab:halfscore_results} shows the our final HalFscore includes precision, recall, and fscore, and further classifies hallucinations into object, attribute, and relation types for more detailed evaluation. 
From this metric, we observe that compared to the LLaVA1.5 baseline, the leading methods have achieved substantial and comprehensive improvements in both precision and recall. 
Notably, there is a significant reduction in hallucinations when describing objects and attributes, although there remains room for improvement in relation hallucinations. 
Our method also demonstrates marked improvement over LLaVA1.5, with a +6.2 increase in precision, +0.7 in recall, and +2.4 in Fscore. In comparisons with OPERA, VCD, and RLAIF-V, our approach achieved the best results in precision and fscore, though the improvement in recall is less pronounced. 
Overall, by leveraging these hallucination mitigation techniques, OPERA, VCD, and RLAIF-V have achieved recall scores comparable to MiniCPM-2.6, while our method has even surpassed MiniCPM, further underscoring the effectiveness of our approach.

\paragraph{CHAIR Evaluation} 
To evaluate caption-based object hallucination, we use the CHAIR metric \citep{rohrbach2018object}, which measures the percentage of generated words that correspond to actual objects in the image. 
The results in Table \ref{tab:main_results} show that, on this metric, our method significantly reduces hallucinations compared to the LLaVA1.5 baseline and outperforms both OPERA and VCD by a large margin. 
However, RLAIF-V performs better on this metric, showing fewer object hallucinations. 
A possible reason is that it uses additional feedback model. 
Moving forward, we plan to design targeted perturbation texts for objects to further enhance performance. 

\paragraph{HallusionBench Evaluation} 
HallusionBench \citep{guan2024hallusionbench} evaluates how well multi-modal models handle language hallucinations and visual illusions.
Previous hallucination assessment mainly focused on dense caption tasks, and by incorporating HallusionBench, we aim to enhance our hallucination evaluation through the addition of multiple-choice and vision reasoning tasks. 
The results in Table \ref{tab:main_results} demonstrate a substantial improvement (+0.6) of our method on the HallusionBench metric, whereas the VCD decoding strategy merely maintained the baseline, OPERA led to a modest increase of 0.2. 
This outcome further confirms that our method is not only effective in dense captioning tasks but also excels in vision reasoning tasks.

\paragraph{General Multimodal Ability} 
We assess the models' general ability on three widely used benchmarks: MMBench \citep{liu2023mmbench}, CCBench \citep{liu2023mmbench}, and SEEDImage \citep{li2023seed}, to verify that our model's general capabilities.
From the Table \ref{tab:main_results}, we observe that both VCD and RLAIF-V result in some degradation in general performance, while OPERA remains stable. 
In contrast, our method not only avoids degradation in general metrics but also shows improvements across multiple areas, with gains of +1.6 in MMB, +0.3 in SEEDImage, and +1.2 in CCBench.
 We attribute this improvement in generalization to perturbation training, which compels the multimodal model to more effectively leverage image information.

\subsection{Analyses}

\begin{table}[t]
\centering
\renewcommand{\arraystretch}{1.5} 
\small 
\scalebox{0.9}
{
\begin{tabular}{ p{2cm}|p{11.3cm} } 
\hline
\textbf{Type}                    & \textbf{Example}                                                                                                                             \\ \hline

\textbf{Version1}      &  \texttt{<image>} What follows presents incorrect guidance based on the image. \texttt{<perturbation text>} Please ignore any false guidance and focus on the image to answer the question below. \texttt{<Question>}                                                                                  \\ \hline
\textbf{Version2}  & \texttt{<image>} \texttt{<perturbation text>} Please examine the image and respond to the question below. \texttt{<Question>}          \\ \hline
\textbf{Version3}  &
\texttt{<image>} \texttt{<perturbation text>}  \texttt{<Question>}                                                                                       \\ \hline

\textbf{Random}  &
\texttt{<image>} Coming up is a portion of non-visual text. \texttt{<random perturbation text>} Ignore the previous perturbation text and focus on the image to complete the task. \texttt{<Question>}                                                                                       \\ \hline

\end{tabular}
}
\caption{Different variants of perturbative visual training.}
\label{tab:variants}
\end{table}

\begin{table}[t]
\renewcommand{\arraystretch}{1.4}
\setlength\tabcolsep{1.2pt} 
\centering
\footnotesize
{
\scalebox{0.9}
{
\begin{tabular}{l  ccc ccc c ccc}
\toprule
\multirow{2}{*}{\textbf{Type}} 
& \multicolumn{3}{c}{\textbf{HalFscore}$\uparrow$} 
& \multicolumn{3}{c}{\textbf{Object HalBench}$\downarrow$} 
& \multirow{2}{*}{\textbf{HalBench}$\uparrow$} 
& \multirow{2}{*}{\textbf{MMB}$\uparrow$} 
& \multirow{2}{*}{\textbf{SEED}$\uparrow$} 
& \multirow{2}{*}{\textbf{CCBench}$\uparrow$} \\
\cmidrule(lr){2-4} \cmidrule(lr){5-7}
& \textbf{Precision} & \textbf{Recall} & \textbf{Fscore} 
& \textbf{CHAIR$_{s}$} & \textbf{CHAIR$_{i}$} & \textbf{length} & & & & \\
\midrule
LLaVA1.5  & 53.3 & 45.8 & 49.2 & 54.2 & 15.0 & 100 & 46.9 & 67.3 & 65.3 & 29.4 \\
\midrule
Version1 & 59.5 & \underline{46.5} & \underline{52.2} & 36.1 & 10.4 & 89 & 47.5 & \underline{68.9} & \underline{65.6} & \underline{30.6} \\
Version2 & 59.6 & 46.0 & 52.0 & 33.6 & \underline{10.2} & 82 & 49.2 & 67.8 & 65 & 29.2  \\
Version3 & \underline{59.7} & 46.1 & 52.0 & \underline{32.3} & 10.6 & 78 & \underline{49.6} & 66.5 & 64.7 & 28.8  \\
\midrule
Random   & 57.1 & 45.6 & 50.7 & 52.4 & 15.0 & 99 & 49.0 & 67.4 & 65.1 & 29.4 \\
\bottomrule
\end{tabular}
}
}
\caption{Effect of varying perturbation levels during training.The best results are underlined. 
}
\label{tab:ablation}
\end{table}

\paragraph{Impact of the Perturbation Strengths} To assess the effect of perturbation levels on training, we devised three methods of inserting perturbation text, detailed in Table \ref{tab:variants}. The first method alerts the model to upcoming text perturbations, instructing it to ignore them. The second focuses the model on image regions without referencing perturbations. The third offers no guidance. These methods progressively increase the level of perturbation and the difficulty of training the multimodal model. As Table \ref{tab:ablation} shows, increasing perturbation levels enhances training difficulty and reduces hallucinations, as indicated by HalFscore and CHAIR metrics. However, this also leads to shorter captions and lower recall, suggesting a more cautious model behavior. General performance metrics decline consistently, implying that while perturbations decrease hallucinations, excessive perturbation use may impair overall performance.

\paragraph{Impact of the Perturbation Relevance} To investigate the impact of relevance of perturbation text, we conducted experiments using randomly sampled texts from a text dataset instead of the carefully designed perturbations used earlier. 
The results in Table \ref{tab:ablation} show that introducing random text perturbations, compared to the LLaVA1.5 baseline, still mitigates hallucinations in multimodal models, as evidenced by improvements in HaFscore and HalBench. However, the effectiveness of random perturbations is inferior to that of targeted text-based experiments. We suggest that even random perturbations influence the training of multimodal models, as the training is disrupted by the random text interference. However, more relevant perturbations exert a stronger disruptive effect, prompting the model to focus more on the image to overcome these challenges.

\begin{table}[t]
\renewcommand{\arraystretch}{1.2}
\centering
\footnotesize
\scalebox{0.9}
{
\begin{tabular}{ r r c }
\toprule
\textbf{Auto Evaluation}  &  \textbf{Human Evaluation}  & \textbf{Pearson Correlation}   \\
\midrule
Recall   & Recall\_h  &  $78.1$ \\ 
Precision       & Precision\_h  & $80.7$ \\
MMhalBench     & Precision\_h  & $71.7$ \\ 
\bottomrule
\end{tabular}
}
\caption{Correlations between automatically computed metrics and human subjective opinions.}
\label{tab:user_study}
\end{table}

\paragraph{Human Analysis of HalFscore} 

We perform the user study to explore whether the proposed HalFscore correlates well with human evaluation. 
We show an image and ask human raters to compare two paragraphs generated 
and determine which generated text describes the main subject and detailed information of the input image more comprehensively (recall\_h), and which text contains less information that is not present in the input image (precision\_h). 
Four methods, LLaVA 1.5, MiniCPM-2.6, Internvl2, and our method are selected for human evaluation. 
Each method is compared with other methods for $12$ times. 
We compute the human evaluated scores for the recall and precision and drive the Pearson correlations coefficients between the scores of human evaluation and that of HalFscore. 
The results are shown in the Table \ref{tab:user_study}. 
We observe that our precision score and recall score align with human evaluations.
Specifically, we compare our method with another metric, MMhalbench. 
The correlation coefficient between the metric and the human evaluations of precision scores is relatively smaller than HalFscore.
The comparison demonstrates our metric aligns with human evaluations more than directly asking large language models to rate the level of hallucination. 
Further details about the user study can be found in Section \ref{appendix:user_study} of the Appendix. 


\paragraph{Complementary Effect to Decoding Strategies.} 
PerturboLLaVA is a novel training strategy that enhances model performance in the SFT phase. 
It can be integrated seamlessly with post-SFT optimization RLAIF-V and OPERA during inference. In our tests, using OPERA as a plugin with PerturboLLaVA during inference led to performance boosts, as detailed in Table \ref{tab:main_results}. PerturboLLaVA with OPERA achieved further gains, improving HalFscore by 0.6, with a 3-point improvement in the CHAIR\textsubscript{s}. Hallucination reasoning in Hallusionbench also improved by +0.1, while general capability increased by 0.4 in CCBench. 
Compared to applying OPERA on LLaVA1.5, replacing the baseline model with PerturboLLaVA resulted in improved performance in both hallucination reduction and general capability. 
Thus, PerturboLLaVA introduces a new optimization direction that complements existing strategies, offering an additional benefit. 

\section{Conclusion}
In our work, we introduce a concept-level HalFscore that enables fine-grained analysis of various hallucinations involving objects, attributes, and relations in dense captions. 
This metric also reflects the overall captioning capability of multi-modal models. To mitigate hallucinations, we propose a simple yet effective training strategy that guides multi-modal models to reduce reliance on linguistic priors and prioritize image information. Our method outperforms leading state-of-the-art approaches without incurring additional training or inference overhead. Overall, we believe that our proposed metric enhances the evaluation of hallucinations in multi-modal models and that our method effectively mitigates hallucination issues. The proposed method shows great promise of becoming a standard strategy for training robust multimodal models.

\section*{Acknowledgments}
This work was supported by the National Key R\&D Program of China (No.\ 2022ZD0160101), the Ningbo Science and Technology Bureau 
(Grant Number 2024Z291), and the National Natural Science Foundation of China (No. 62206244).
CC and ML contributed equally. Part of the work
was done when CC was doing an internship at WeChat Group.

\bibliography{main}

\begin{thebibliography}{45}
\providecommand{\natexlab}[1]{#1}
\providecommand{\url}[1]{\texttt{#1}}
\expandafter\ifx\csname urlstyle\endcsname\relax
  \providecommand{\doi}[1]{doi: #1}\else
  \providecommand{\doi}{doi: \begingroup \urlstyle{rm}\Url}\fi

\bibitem[Achiam et~al.(2023)Achiam, Adler, Agarwal, Ahmad, Akkaya, Aleman, Almeida, Altenschmidt, Altman, Anadkat, et~al.]{achiam2023gpt}
Josh Achiam, Steven Adler, Sandhini Agarwal, Lama Ahmad, Ilge Akkaya, Florencia~Leoni Aleman, Diogo Almeida, Janko Altenschmidt, Sam Altman, Shyamal Anadkat, et~al.
\newblock Gpt-4 technical report.
\newblock \emph{arXiv preprint arXiv:2303.08774}, 2023.

\bibitem[Bai et~al.(2023)Bai, Bai, Yang, Wang, Tan, Wang, Lin, Zhou, and Zhou]{bai2023qwen}
Jinze Bai, Shuai Bai, Shusheng Yang, Shijie Wang, Sinan Tan, Peng Wang, Junyang Lin, Chang Zhou, and Jingren Zhou.
\newblock Qwen-vl: A frontier large vision-language model with versatile abilities.
\newblock \emph{arXiv preprint arXiv:2308.12966}, 2023.

\bibitem[Bi et~al.(2024)Bi, Chen, Chen, Chen, Dai, Deng, Ding, Dong, Du, Fu, et~al.]{bi2024deepseek}
Xiao Bi, Deli Chen, Guanting Chen, Shanhuang Chen, Damai Dai, Chengqi Deng, Honghui Ding, Kai Dong, Qiushi Du, Zhe Fu, et~al.
\newblock Deepseek llm: Scaling open-source language models with longtermism.
\newblock \emph{arXiv preprint arXiv:2401.02954}, 2024.

\bibitem[Bradley \& Terry(1952)Bradley and Terry]{bradley1952rank}
Ralph~Allan Bradley and Milton~E Terry.
\newblock Rank analysis of incomplete block designs: I. the method of paired comparisons.
\newblock \emph{Biometrika}, 39\penalty0 (3/4):\penalty0 324--345, 1952.

\bibitem[Chen et~al.(2023)Chen, Li, Dong, Zhang, He, Wang, Zhao, and Lin]{chen2023sharegpt4v}
Lin Chen, Jinsong Li, Xiaoyi Dong, Pan Zhang, Conghui He, Jiaqi Wang, Feng Zhao, and Dahua Lin.
\newblock Sharegpt4v: Improving large multi-modal models with better captions.
\newblock \emph{arXiv preprint arXiv: 2311.12793}, 2023.

\bibitem[Chen et~al.(2024)Chen, Wu, Wang, Su, Chen, Xing, Zhong, Zhang, Zhu, Lu, Li, Luo, Lu, Qiao, and Dai]{chen2023internvl}
Zhe Chen, Jiannan Wu, Wenhai Wang, Weijie Su, Guo Chen, Sen Xing, Muyan Zhong, Qinglong Zhang, Xizhou Zhu, Lewei Lu, Bin Li, Ping Luo, Tong Lu, Yu~Qiao, and Jifeng Dai.
\newblock Internvl: Scaling up vision foundation models and aligning for generic visual-linguistic tasks.
\newblock \emph{CVPR}, 2024.

\bibitem[Chiang et~al.(2023)Chiang, Li, Lin, Sheng, Wu, Zhang, Zheng, Zhuang, Zhuang, Gonzalez, et~al.]{chiang2023vicuna}
Wei-Lin Chiang, Zhuohan Li, Zi~Lin, Ying Sheng, Zhanghao Wu, Hao Zhang, Lianmin Zheng, Siyuan Zhuang, Yonghao Zhuang, Joseph~E Gonzalez, et~al.
\newblock Vicuna: An open-source chatbot impressing gpt-4 with 90\%* chatgpt quality.
\newblock \emph{See https://vicuna. lmsys. org (accessed 14 April 2023)}, 2023.

\bibitem[Clark et~al.(2019)Clark, Yatskar, and Zettlemoyer]{clark2019don}
Christopher Clark, Mark Yatskar, and Luke Zettlemoyer.
\newblock Don’t take the easy way out: Ensemble based methods for avoiding known dataset biases.
\newblock In \emph{Proceedings of the 2019 Conference on Empirical Methods in Natural Language Processing and the 9th International Joint Conference on Natural Language Processing (EMNLP-IJCNLP)}, pp.\  4069--4082, 2019.

\bibitem[Dai et~al.(2023)Dai, Li, Li, Tiong, Zhao, Wang, Li, Fung, and Hoi]{instructblip}
Wenliang Dai, Junnan Li, Dongxu Li, Anthony Meng~Huat Tiong, Junqi Zhao, Weisheng Wang, Boyang Li, Pascale Fung, and Steven Hoi.
\newblock Instructblip: Towards general-purpose vision-language models with instruction tuning, 2023.

\bibitem[Dubey et~al.(2024)Dubey, Jauhri, Pandey, Kadian, Al-Dahle, Letman, Mathur, Schelten, Yang, Fan, et~al.]{dubey2024llama}
Abhimanyu Dubey, Abhinav Jauhri, Abhinav Pandey, Abhishek Kadian, Ahmad Al-Dahle, Aiesha Letman, Akhil Mathur, Alan Schelten, Amy Yang, Angela Fan, et~al.
\newblock The llama 3 herd of models.
\newblock \emph{arXiv preprint arXiv:2407.21783}, 2024.

\bibitem[Guan et~al.(2024)Guan, Liu, Wu, Xian, Li, Liu, Wang, Chen, Huang, Yacoob, et~al.]{guan2024hallusionbench}
Tianrui Guan, Fuxiao Liu, Xiyang Wu, Ruiqi Xian, Zongxia Li, Xiaoyu Liu, Xijun Wang, Lichang Chen, Furong Huang, Yaser Yacoob, et~al.
\newblock Hallusionbench: an advanced diagnostic suite for entangled language hallucination and visual illusion in large vision-language models.
\newblock In \emph{Proceedings of the IEEE/CVF Conference on Computer Vision and Pattern Recognition}, pp.\  14375--14385, 2024.

\bibitem[Hu et~al.(2024)Hu, Tu, Han, He, Cui, Long, Zheng, Fang, Huang, Zhao, et~al.]{hu2024minicpm}
Shengding Hu, Yuge Tu, Xu~Han, Chaoqun He, Ganqu Cui, Xiang Long, Zhi Zheng, Yewei Fang, Yuxiang Huang, Weilin Zhao, et~al.
\newblock Minicpm: Unveiling the potential of small language models with scalable training strategies.
\newblock \emph{arXiv preprint arXiv:2404.06395}, 2024.

\bibitem[Huang et~al.(2024)Huang, Dong, Zhang, Wang, He, Wang, Lin, Zhang, and Yu]{huang2024opera}
Qidong Huang, Xiaoyi Dong, Pan Zhang, Bin Wang, Conghui He, Jiaqi Wang, Dahua Lin, Weiming Zhang, and Nenghai Yu.
\newblock Opera: Alleviating hallucination in multi-modal large language models via over-trust penalty and retrospection-allocation.
\newblock In \emph{Proceedings of the IEEE/CVF Conference on Computer Vision and Pattern Recognition}, pp.\  13418--13427, 2024.

\bibitem[Jing et~al.(2020)Jing, Wu, Zhang, Yunde, and Wu]{jing2020overcoming}
Chenchen Jing, Yuwei Wu, Xiaoxun Zhang, Jia Yunde, and Qi~Wu.
\newblock Overcoming language priors in vqa via decomposed linguistic representations.
\newblock In \emph{Thirty-Forth AAAI Conference on Artificial Intelligence (AAAI)}, pp.\  11181--11188, 2020.

\bibitem[Lauren{\c{c}}on et~al.(2024)Lauren{\c{c}}on, Marafioti, Sanh, and Tronchon]{idefics3}
Hugo Lauren{\c{c}}on, Andr{\'e}s Marafioti, Victor Sanh, and L{\'e}o Tronchon.
\newblock Building and better understanding vision-language models: insights and future directions.
\newblock \emph{arXiv preprint arXiv:2408.12637}, 2024.

\bibitem[Leng et~al.(2024)Leng, Zhang, Chen, Li, Lu, Miao, and Bing]{VCD}
Sicong Leng, Hang Zhang, Guanzheng Chen, Xin Li, Shijian Lu, Chunyan Miao, and Lidong Bing.
\newblock Mitigating object hallucinations in large vision-language models through visual contrastive decoding.
\newblock In \emph{Proceedings of the IEEE/CVF Conference on Computer Vision and Pattern Recognition}, pp.\  13872--13882, 2024.

\bibitem[Li et~al.(2024)Li, Zhang, Guo, Zhang, Li, Zhang, Zhang, Li, Liu, and Li]{li2024llava}
Bo~Li, Yuanhan Zhang, Dong Guo, Renrui Zhang, Feng Li, Hao Zhang, Kaichen Zhang, Yanwei Li, Ziwei Liu, and Chunyuan Li.
\newblock Llava-onevision: Easy visual task transfer.
\newblock \emph{arXiv preprint arXiv:2408.03326}, 2024.

\bibitem[Li et~al.(2023{\natexlab{a}})Li, Wang, Wang, Ge, Ge, and Shan]{li2023seed}
Bohao Li, Rui Wang, Guangzhi Wang, Yuying Ge, Yixiao Ge, and Ying Shan.
\newblock Seed-bench: Benchmarking multimodal llms with generative comprehension.
\newblock \emph{arXiv preprint arXiv:2307.16125}, 2023{\natexlab{a}}.

\bibitem[Li et~al.(2023{\natexlab{b}})Li, Du, Zhou, Wang, Zhao, and Wen]{li2023evaluating}
Yifan Li, Yifan Du, Kun Zhou, Jinpeng Wang, Wayne~Xin Zhao, and Ji-Rong Wen.
\newblock Evaluating object hallucination in large vision-language models.
\newblock \emph{arXiv preprint arXiv:2305.10355}, 2023{\natexlab{b}}.

\bibitem[Liu et~al.(2023{\natexlab{a}})Liu, Lin, Li, Wang, Yacoob, and Wang]{liu2023aligning}
Fuxiao Liu, Kevin Lin, Linjie Li, Jianfeng Wang, Yaser Yacoob, and Lijuan Wang.
\newblock Aligning large multi-modal model with robust instruction tuning.
\newblock \emph{arXiv preprint arXiv:2306.14565}, 2023{\natexlab{a}}.

\bibitem[Liu et~al.(2023{\natexlab{b}})Liu, Lin, Li, Wang, Yacoob, and Wang]{liu2023mitigating}
Fuxiao Liu, Kevin Lin, Linjie Li, Jianfeng Wang, Yaser Yacoob, and Lijuan Wang.
\newblock Mitigating hallucination in large multi-modal models via robust instruction tuning.
\newblock In \emph{The Twelfth International Conference on Learning Representations}, 2023{\natexlab{b}}.

\bibitem[Liu et~al.(2024{\natexlab{a}})Liu, Li, Li, Li, Zhang, Shen, and Lee]{liu2024llavanext}
Haotian Liu, Chunyuan Li, Yuheng Li, Bo~Li, Yuanhan Zhang, Sheng Shen, and Yong~Jae Lee.
\newblock Llava-next: Improved reasoning, ocr, and world knowledge, January 2024{\natexlab{a}}.
\newblock URL \url{https://llava-vl.github.io/blog/2024-01-30-llava-next/}.

\bibitem[Liu et~al.(2024{\natexlab{b}})Liu, Li, Wu, and Lee]{llava}
Haotian Liu, Chunyuan Li, Qingyang Wu, and Yong~Jae Lee.
\newblock Visual instruction tuning.
\newblock \emph{Advances in neural information processing systems}, 36, 2024{\natexlab{b}}.

\bibitem[Liu et~al.(2023{\natexlab{c}})Liu, Duan, Zhang, Li, Zhang, Zhao, Yuan, Wang, He, Liu, et~al.]{liu2023mmbench}
Yuan Liu, Haodong Duan, Yuanhan Zhang, Bo~Li, Songyang Zhang, Wangbo Zhao, Yike Yuan, Jiaqi Wang, Conghui He, Ziwei Liu, et~al.
\newblock Mmbench: Is your multi-modal model an all-around player?
\newblock \emph{arXiv preprint arXiv:2307.06281}, 2023{\natexlab{c}}.

\bibitem[Lu et~al.(2024{\natexlab{a}})Lu, Liu, Zhang, Wang, Dong, Liu, Sun, Ren, Li, Sun, et~al.]{lu2024deepseek}
Haoyu Lu, Wen Liu, Bo~Zhang, Bingxuan Wang, Kai Dong, Bo~Liu, Jingxiang Sun, Tongzheng Ren, Zhuoshu Li, Yaofeng Sun, et~al.
\newblock Deepseek-vl: towards real-world vision-language understanding.
\newblock \emph{arXiv preprint arXiv:2403.05525}, 2024{\natexlab{a}}.

\bibitem[Lu et~al.(2024{\natexlab{b}})Lu, Li, Chen, Xu, Luo, Zhang, and Ye]{lu2024ovis}
Shiyin Lu, Yang Li, Qing-Guo Chen, Zhao Xu, Weihua Luo, Kaifu Zhang, and Han-Jia Ye.
\newblock Ovis: Structural embedding alignment for multimodal large language model.
\newblock \emph{arXiv preprint arXiv:2405.20797}, 2024{\natexlab{b}}.

\bibitem[OpenGVLab(2024)]{internvl2_2024}
OpenGVLab.
\newblock Internvl2: Better than the best—expanding performance boundaries of open-source multimodal models with the progressive scaling strategy, 2024.
\newblock URL \url{https://internvl.github.io/blog/2024-07-02-InternVL-2.0/}.

\bibitem[Ouyang et~al.(2022)Ouyang, Wu, Jiang, Almeida, Wainwright, Mishkin, Zhang, Agarwal, Slama, Ray, et~al.]{ouyang2022training}
Long Ouyang, Jeffrey Wu, Xu~Jiang, Diogo Almeida, Carroll Wainwright, Pamela Mishkin, Chong Zhang, Sandhini Agarwal, Katarina Slama, Alex Ray, et~al.
\newblock Training language models to follow instructions with human feedback.
\newblock \emph{Advances in neural information processing systems}, 35:\penalty0 27730--27744, 2022.

\bibitem[Rafailov et~al.(2024)Rafailov, Sharma, Mitchell, Manning, Ermon, and Finn]{rafailov2024direct}
Rafael Rafailov, Archit Sharma, Eric Mitchell, Christopher~D Manning, Stefano Ermon, and Chelsea Finn.
\newblock Direct preference optimization: Your language model is secretly a reward model.
\newblock \emph{Advances in Neural Information Processing Systems}, 36, 2024.

\bibitem[Rohrbach et~al.(2018)Rohrbach, Hendricks, Burns, Darrell, and Saenko]{rohrbach2018object}
Anna Rohrbach, Lisa~Anne Hendricks, Kaylee Burns, Trevor Darrell, and Kate Saenko.
\newblock Object hallucination in image captioning.
\newblock In \emph{Proceedings of the 2018 Conference on Empirical Methods in Natural Language Processing}, pp.\  4035--4045, 2018.

\bibitem[Sun et~al.(2023)Sun, Shen, Cao, Liu, Li, Shen, Gan, Gui, Wang, Yang, et~al.]{sun2023aligning}
Zhiqing Sun, Sheng Shen, Shengcao Cao, Haotian Liu, Chunyuan Li, Yikang Shen, Chuang Gan, Liang-Yan Gui, Yu-Xiong Wang, Yiming Yang, et~al.
\newblock Aligning large multimodal models with factually augmented rlhf.
\newblock \emph{arXiv preprint arXiv:2309.14525}, 2023.

\bibitem[Touvron et~al.(2023{\natexlab{a}})Touvron, Lavril, Izacard, Martinet, Lachaux, Lacroix, Rozi{\`e}re, Goyal, Hambro, Azhar, et~al.]{touvron2023llama}
Hugo Touvron, Thibaut Lavril, Gautier Izacard, Xavier Martinet, Marie-Anne Lachaux, Timoth{\'e}e Lacroix, Baptiste Rozi{\`e}re, Naman Goyal, Eric Hambro, Faisal Azhar, et~al.
\newblock Llama: Open and efficient foundation language models.
\newblock \emph{arXiv preprint arXiv:2302.13971}, 2023{\natexlab{a}}.

\bibitem[Touvron et~al.(2023{\natexlab{b}})Touvron, Martin, Stone, Albert, Almahairi, Babaei, Bashlykov, Batra, Bhargava, Bhosale, et~al.]{touvron2023llama2}
Hugo Touvron, Louis Martin, Kevin Stone, Peter Albert, Amjad Almahairi, Yasmine Babaei, Nikolay Bashlykov, Soumya Batra, Prajjwal Bhargava, Shruti Bhosale, et~al.
\newblock Llama 2: Open foundation and fine-tuned chat models.
\newblock \emph{arXiv preprint arXiv:2307.09288}, 2023{\natexlab{b}}.

\bibitem[Urbanek et~al.(2023)Urbanek, Bordes, Astolfi, Williamson, Sharma, and Romero-Soriano]{Urbanek2023API}
Jack Urbanek, Florian Bordes, Pietro Astolfi, Mary Williamson, Vasu Sharma, and Adriana Romero-Soriano.
\newblock A picture is worth more than 77 text tokens: Evaluating clip-style models on dense captions.
\newblock \emph{2024 IEEE/CVF Conference on Computer Vision and Pattern Recognition (CVPR)}, pp.\  26690--26699, 2023.
\newblock URL \url{https://api.semanticscholar.org/CorpusID:266209761}.

\bibitem[Wang et~al.(2023{\natexlab{a}})Wang, Wu, Han, Peng, Zhong, Zhang, Dong, Li, Li, Wang, et~al.]{wang2023vigc}
Bin Wang, Fan Wu, Xiao Han, Jiahui Peng, Huaping Zhong, Pan Zhang, Xiaoyi Dong, Weijia Li, Wei Li, Jiaqi Wang, et~al.
\newblock Vigc: Visual instruction generation and correction.
\newblock \emph{arXiv preprint arXiv:2308.12714}, 2023{\natexlab{a}}.

\bibitem[Wang et~al.(2023{\natexlab{b}})Wang, Wang, Xu, Zhang, Gu, Jia, Yan, Zhang, and Sang]{wang2023llm}
Junyang Wang, Yuhang Wang, Guohai Xu, Jing Zhang, Yukai Gu, Haitao Jia, Ming Yan, Ji~Zhang, and Jitao Sang.
\newblock An llm-free multi-dimensional benchmark for mllms hallucination evaluation.
\newblock \emph{arXiv preprint arXiv:2311.07397}, 2023{\natexlab{b}}.

\bibitem[Wang et~al.(2024)Wang, Bai, Tan, Wang, Fan, Bai, Chen, Liu, Wang, Ge, et~al.]{wang2024qwen2}
Peng Wang, Shuai Bai, Sinan Tan, Shijie Wang, Zhihao Fan, Jinze Bai, Keqin Chen, Xuejing Liu, Jialin Wang, Wenbin Ge, et~al.
\newblock Qwen2-vl: Enhancing vision-language model's perception of the world at any resolution.
\newblock \emph{arXiv preprint arXiv:2409.12191}, 2024.

\bibitem[Wu et~al.(2017)Wu, Teney, Wang, Shen, Dick, and van~den Hengel]{wu2017visual}
Qi~Wu, Damien Teney, Peng Wang, Chunhua Shen, Anthony Dick, and Anton van~den Hengel.
\newblock Visual question answering: A survey of methods and datasets.
\newblock \emph{Computer Vision and Image Understanding}, 163:\penalty0 21--40, 2017.

\bibitem[Yang et~al.(2024)Yang, Yang, Hui, Zheng, Yu, Zhou, Li, Li, Liu, Huang, et~al.]{yang2024qwen2llm}
An~Yang, Baosong Yang, Binyuan Hui, Bo~Zheng, Bowen Yu, Chang Zhou, Chengpeng Li, Chengyuan Li, Dayiheng Liu, Fei Huang, et~al.
\newblock Qwen2 technical report.
\newblock \emph{arXiv preprint arXiv:2407.10671}, 2024.

\bibitem[Yin et~al.(2023)Yin, Fu, Zhao, Xu, Wang, Sui, Shen, Li, Sun, and Chen]{yin2023woodpecker}
Shukang Yin, Chaoyou Fu, Sirui Zhao, Tong Xu, Hao Wang, Dianbo Sui, Yunhang Shen, Ke~Li, Xing Sun, and Enhong Chen.
\newblock Woodpecker: Hallucination correction for multimodal large language models.
\newblock \emph{arXiv preprint arXiv:2310.16045}, 2023.

\bibitem[Yu et~al.(2024)Yu, Zhang, Yao, Dang, Chen, Lu, Cui, He, Liu, Chua, et~al.]{yu2024rlaif}
Tianyu Yu, Haoye Zhang, Yuan Yao, Yunkai Dang, Da~Chen, Xiaoman Lu, Ganqu Cui, Taiwen He, Zhiyuan Liu, Tat-Seng Chua, et~al.
\newblock Rlaif-v: Aligning mllms through open-source ai feedback for super gpt-4v trustworthiness.
\newblock \emph{arXiv e-prints}, pp.\  arXiv--2405, 2024.

\bibitem[Zhang et~al.(2023)Zhang, Wang, Cao, Xu, Ouyang, Zhao, Ding, Zhang, Duan, Yan, et~al.]{zhang2023internlm}
Pan Zhang, Xiaoyi Dong~Bin Wang, Yuhang Cao, Chao Xu, Linke Ouyang, Zhiyuan Zhao, Shuangrui Ding, Songyang Zhang, Haodong Duan, Hang Yan, et~al.
\newblock Internlm-xcomposer: A vision-language large model for advanced text-image comprehension and composition.
\newblock \emph{arXiv preprint arXiv:2309.15112}, 2023.

\bibitem[Zhou et~al.(2024)Zhou, Shu, Zhao, Wu, Xiao, Yang, Xiong, Zhang, Huang, and Liu]{zhou2024mlvu}
Junjie Zhou, Yan Shu, Bo~Zhao, Boya Wu, Shitao Xiao, Xi~Yang, Yongping Xiong, Bo~Zhang, Tiejun Huang, and Zheng Liu.
\newblock Mlvu: A comprehensive benchmark for multi-task long video understanding.
\newblock \emph{arXiv preprint arXiv:2406.04264}, 2024.

\bibitem[Zhou et~al.(2023)Zhou, Cui, Yoon, Zhang, Deng, Finn, Bansal, and Yao]{zhou2023analyzing}
Yiyang Zhou, Chenhang Cui, Jaehong Yoon, Linjun Zhang, Zhun Deng, Chelsea Finn, Mohit Bansal, and Huaxiu Yao.
\newblock Analyzing and mitigating object hallucination in large vision-language models.
\newblock \emph{arXiv preprint arXiv:2310.00754}, 2023.

\bibitem[Zhu et~al.(2023)Zhu, Chen, Shen, Li, and Elhoseiny]{zhu2023minigpt}
Deyao Zhu, Jun Chen, Xiaoqian Shen, Xiang Li, and Mohamed Elhoseiny.
\newblock Minigpt-4: Enhancing vision-language understanding with advanced large language models.
\newblock \emph{arXiv preprint arXiv:2304.10592}, 2023.

\end{thebibliography}
\bibliographystyle{main}

\newpage

\appendix
\section{Appendix}

\subsection{Implementation details of HalFscore}
\subsubsection{statistic of image-caption pair}

\begin{table}[h]
\renewcommand{\arraystretch}{1.2}
\footnotesize 
\centering
\begin{minipage}{0.55\textwidth} 
\centering
{
\begin{tabular}{l r c c }
\toprule
\textbf{Statistics} & \textbf{Number} & \textbf{Ratio} \\
\midrule
Indoor Scene & 307 & 30.7\% \\
  \quad Public Spaces & 134 & 13.4\% \\
  \quad Home & 79 & 7.9\% \\
  \quad Stores & 49 & 4.9\% \\
  \quad Office & 27 & 2.7\% \\
  \quad Others & 18 & 1.8\% \\
\midrule
Outdoor Scene & 693 & 69.3\% \\
  \quad Urban & 194 & 19.4\% \\
  \quad Architecture & 145 & 14.5\% \\
  \quad Transportation & 152 & 15.2\% \\
  \quad Natural Scenery & 128 & 12.8\% \\
  \quad Rural & 68 & 6.8\% \\
  \quad Others & 6 & 0.6\% \\
\midrule
Total Number of Image & 1,000&100\% \\
\bottomrule
\end{tabular}
}
\caption{Statistic of our image data.}
\label{tab:halfscore_results2}
\end{minipage}%
\hfill
\begin{minipage}{0.4\textwidth} 
\centering
\includegraphics[width=\linewidth]{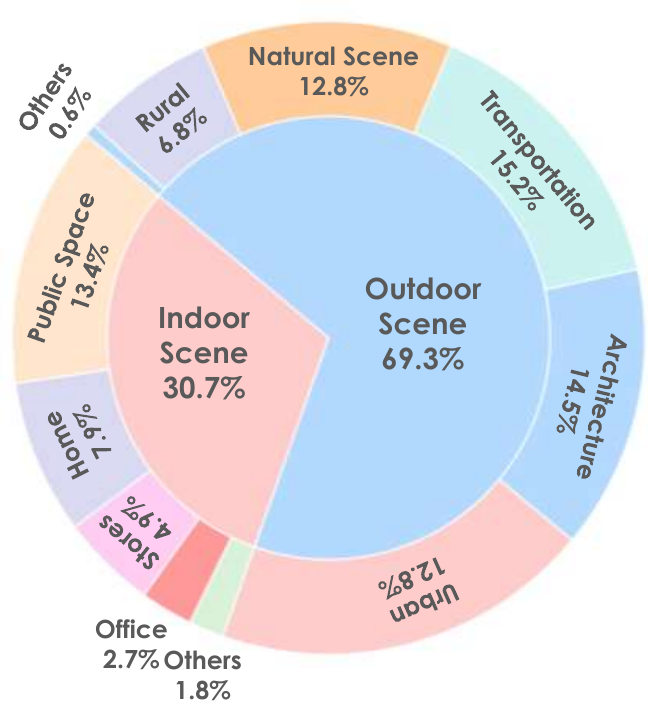}
\caption{Distribution of our image data.}
\end{minipage}
\end{table}

We manually selected 1,000 images from the DCI dataset to ensure that the final image data used is characterized by high quality and diversity. The images we ultimately used consist of approximately 30.7\% indoor scenes and 69.3\% outdoor scenes, and we further subdivided both indoor and outdoor scenes into more detailed categories. The statistical results are shown in the figure and table. By using diverse image data, we ensure comprehensive evaluation of the multimodal model's hallucinatory outputs across various scenarios.

\subsubsection{Construction of concept graph}

\begin{figure}[h]
    \centering
    \includegraphics[width=1.0\textwidth]{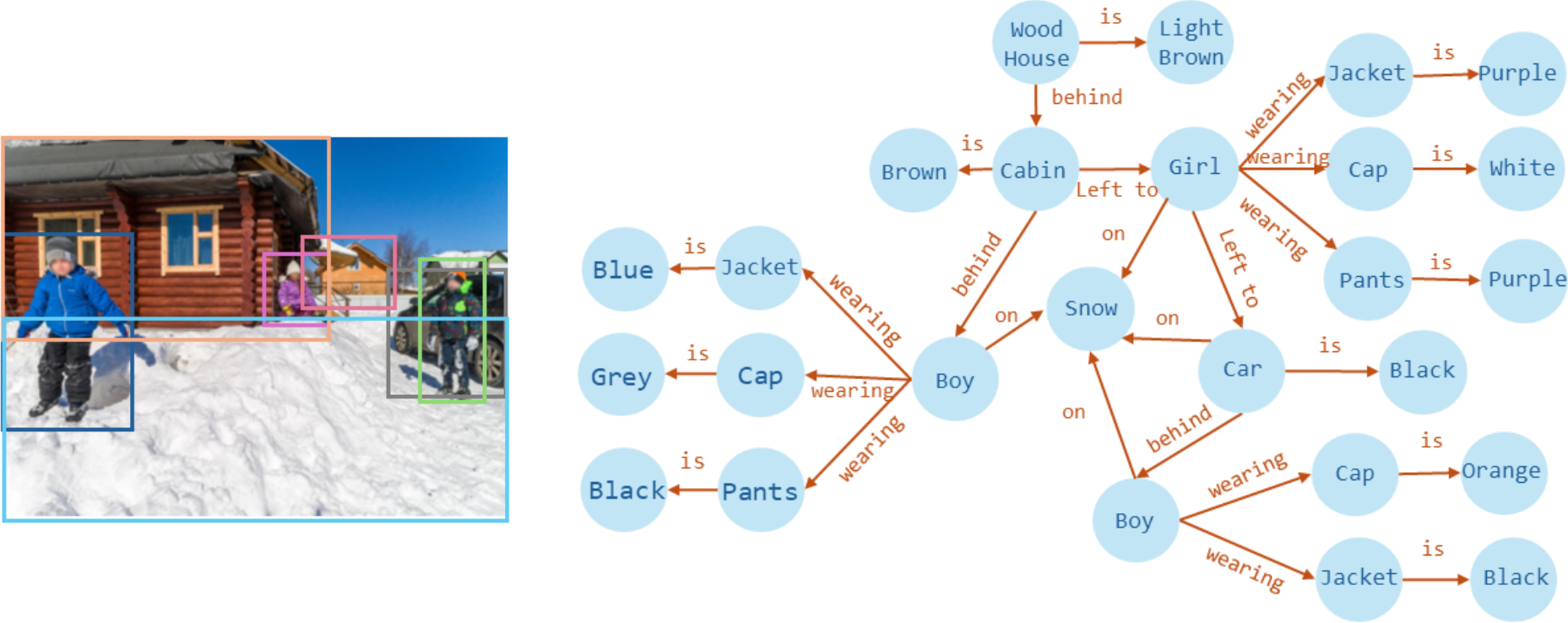} 
    \caption{Example of the construction of concept graph.}
\end{figure}

\begin{figure*}[!th]
\begin{tcolorbox}[
    colback=white, 
    colframe=gray!50!black, 
    coltitle=black, 
    title=\textbf{VLM Caption}, 
    fonttitle=\bfseries\large, 
    arc=4mm, 
    enhanced, 
    attach boxed title to top left={yshift=-\tcboxedtitleheight/2, xshift=10pt}, 
    boxed title style={
        enhanced,
        colback=white,
        colframe=white,
        arc=0mm,
        left=0pt,
        right=0pt,
        boxsep=0pt
    }
]

{
Three children are playing in the snow in front of a log cabin. One child is wearing a blue jacket, another is wearing a pink jacket, and the third child is wearing an orange jacket. They are all wearing hats, with one child wearing an orange hat and the other two wearing blue hats.There is a car parked in front of the cabin, and the children seem to be enjoying themselves as they play in the snow.}
\caption{VLM Caption}
\label{fig: vlm caption}

\end{tcolorbox}
\end{figure*}

\begin{figure*}[!th]
\begin{tcolorbox}[
    colback=white, 
    colframe=gray!50!black, 
    coltitle=black, 
    title=\textbf{VLM Triplets}, 
    fonttitle=\bfseries\large, 
    arc=4mm, 
    enhanced, 
    attach boxed title to top left={yshift=-\tcboxedtitleheight/2, xshift=10pt}, 
    boxed title style={
        enhanced,
        colback=white,
        colframe=white,
        arc=0mm,
        left=0pt,
        right=0pt,
        boxsep=0pt
    }
]

{

\texttt{1. ("object"||CHILDREN||are||three)}

\texttt{2. ("object"||CHILDREN||on||Snow)}

\texttt{3. ("object"||LOG CABIN||behind||boy)}

\texttt{4. ("object"||CHILD||wearing||blue jacket)}

\texttt{5. ("object"||CHILD||wearing||pink jacket)}

\texttt{6. ("object"||CHILD||wearing||orange jacket)}

\texttt{7. ("object"||HATS||all children||wearing)}

\texttt{8. ("object"||HAT||is||orange)}

\texttt{9. ("object"||HATS||is||blue)}

\texttt{10. ("object"||CAR||parked in front of||cabin)}

\texttt{11. ("object"||CHILDREN||enjoying||themselves)}

\texttt{12. ("relationship"||CHILDREN||SNOW||The children are playing in the snow||9)}

\texttt{14. ("relationship"||CHILD||BLUE JACKET||One child is wearing a blue jacket||8)}

\texttt{15. ("relationship"||CHILD||PINK JACKET||One child is wearing a pink jacket||8)}

\texttt{16. ("relationship"||CHILD||ORANGE JACKET||One child is wearing an orange jacket||8)}

\texttt{17. ("relationship"||CHILDREN||HATS||All children are wearing hats||8)}

\texttt{18. ("relationship"||CHILD||ORANGE HAT||One child is wearing an orange hat||8)}

\texttt{19. ("relationship"||CHILDREN||BLUE HATS||Two children are wearing blue hats||8)}

\texttt{20. ("relationship"||CAR||LOG CABIN||The car is parked in front of the log cabin||7)}

\texttt{21. ("relationship"||CHILDREN||ENJOYING THEMSELVES||The children seem to be enjoying themselves as they play in the snow||9)}

}
\caption{VLM Triplets}
\label{fig: vlm triplets}

\end{tcolorbox}
\end{figure*}

\begin{figure*}[!th]
\begin{tcolorbox}[
    colback=white, 
    colframe=gray!50!black, 
    coltitle=black, 
    title=\textbf{VLM Graph}, 
    fonttitle=\bfseries\large, 
    arc=4mm, 
    enhanced, 
    attach boxed title to top left={yshift=-\tcboxedtitleheight/2, xshift=10pt}, 
    boxed title style={
        enhanced,
        colback=white,
        colframe=white,
        arc=0mm,
        left=0pt,
        right=0pt,
        boxsep=0pt
    }
]

{

\texttt{1.("NODE":CHILDREN||"EDGE":are||"NODE":three)}

\texttt{2.("NODE":CHILDREN||"EDGE":on||"NODE":SNOW)}

\texttt{3.("NODE":LOG CABIN||"EDGE":behind||"NODE":BOY)}

\texttt{4.("NODE":CHILDREN||"NODE":SNOW||"EDGE":The children are playing in the snow||9)}

\texttt{5.("NODE":CHILD||"NODE":BLUE JACKET||"EDGE":One child is wearing a blue jacket||8)}

\texttt{6.("NODE":CHILD||"NODE":PINK JACKET||"EDGE":One child is wearing a pink jacket||8)}

\texttt{7.("NODE":CHILD||"NODE":ORANGE JACKET||"EDGE":One child is wearing an orange jacket||8)}

\texttt{8.("NODE":CHILDREN||"NODE":HATS||"EDGE":All children are wearing hats||8)}

\texttt{9.("NODE":CHILD||"NODE":ORANGE HAT||"EDGE":One child is wearing an orange hat||8)}

\texttt{10.("NODE":CHILDREN||"NODE":BLUE HATS||"EDGE":Two children are wearing blue hats||8)}

\texttt{11.("NODE":CAR||"NODE":LOG CABIN||"EDGE":The car is parked in front of the log cabin||7)}

\texttt{12.("NODE":CHILDREN||"NODE":ENJOYING THEMSELVES||"EDGE":The children seem to be enjoying themselves as they play in the snow||9)}

}

\caption{VLM Graph}
\label{fig: graph}
\end{tcolorbox}
\end{figure*}

We demonstrate how a caption is parsed into a concept graph within our evaluation pipeline. First, we utilize triplets to extract several concept groups from the original caption\ref{fig: vlm caption}, which are organized in the form of subjects, objects, and the relationships between them. These triplets efficiently capture information from the caption and are sufficiently general to represent object relationships\ref{fig: vlm triplets}, such as \texttt{<Boy, behind, Cabin>}, \texttt{<Girl, left to, Car>}, as well as attributes like \texttt{<jacket, is, blue>}, and states or actions such as \texttt{<Girl, Wearing, Pants>}. This demonstrates the robustness of triplet-based information extraction. Once the caption is transformed into triplets, we further match these triplets based on the subject and object\ref{fig: graph}. Identical subjects or objects are represented as nodes in the concept graph, while the relations from the original triplets are transformed into edges within the graph. The final concept graph, as shown in the figure, consists of entity concepts as nodes and relation concepts as directed edges. This graph effectively organizes both concrete and abstract concepts from the caption in a structured format, enabling subsequent hallucination and omission analysis.

\subsubsection{Analysis of hallucination}
Here we provide an example of our generated analysis of halluciation\ref{fig:analy of hall} and omission\ref{fig:analy of omission}.


\begin{figure*}[!th]
\begin{tcolorbox}[
    colback=white, 
    colframe=gray!50!black, 
    coltitle=black, 
    title=\textbf{Halluciation Analysis}, 
    fonttitle=\bfseries\large, 
    arc=4mm, 
    enhanced, 
    attach boxed title to top left={yshift=-\tcboxedtitleheight/2, xshift=10pt}, 
    boxed title style={
        enhanced,
        colback=white,
        colframe=white,
        arc=0mm,
        left=0pt,
        right=0pt,
        boxsep=0pt
    }
]

{
\textbf{Analysis:}

- \textbf{Entry 5 in VLM List:} 

    - The GT List does not mention any child wearing a pink jacket. This counts as one hallucination for the incorrect object "CHILD" with the attribute "wearing pink jacket".
    
- \textbf{Entry 6 in VLM List:} 

    - The GT List does not mention any child wearing an orange jacket. This counts as one hallucination for the incorrect object "CHILD" with the attribute "wearing orange jacket".
    
- \textbf{Entry 8 in VLM List:} 

    - The GT List does not mention any child wearing an orange hat. This counts as one hallucination for the incorrect object "HAT" with the attribute "orange".
    
- \textbf{Entry 9 in VLM List:} 

    - The GT List does not mention any child wearing blue hats. This counts as one hallucination for the incorrect object "HATS" with the attribute "blue".
    
- \textbf{Entry 10 in VLM List:} 

    - The GT List mentions a "DARK-COLORED CAR" along the right side and behind the boy on the right, not in front of the cabin. This counts as one hallucination for the incorrect object "CAR" with the attribute "parked in front of cabin".
    
- \textbf{Entry 11 in VLM List:} 

    - The GT List does not mention the children "enjoying themselves". This counts as one hallucination for the incorrect attribute "enjoying themselves".
    
- \textbf{Entry 4 in VLM List:} 

    - "CHILD" with "PINK JACKET" is an incorrect object already counted. This relationship involves an incorrect object, so it does not count separately.
    
- \textbf{Entry 5 in VLM List:} 

    - "CHILD" with "ORANGE JACKET" is an incorrect object already counted. This relationship involves an incorrect object, so it does not count separately.
    
- \textbf{Entry 7 in VLM List:}

    - "HAT" with "ORANGE" is an incorrect object already counted. This relationship involves an incorrect object, so it does not count separately.
    
- \textbf{Entry 8 in VLM List:} 

    - ``HATS" with ``BLUE" is an incorrect object already counted. This relationship involves an incorrect object, so it does not count separately.
    
- \textbf{Entry 9 in VLM List:} 

    - ``CAR" with "parked in front of cabin" is an incorrect object already counted. This relationship involves an incorrect object, so it does not count separately.
    
- \textbf{Entry 10 in VLM List:} 

    - "CHILDREN" with "enjoying themselves" is an incorrect object already counted. This relationship involves an incorrect object, so it does not count separately.nnIncorrect Serial Numbers: 5, 6, 8, 9, 10, 11
    }
\caption{Analysis of Halluciation}
\label{fig:analy of hall}

\end{tcolorbox}
\end{figure*}

\begin{figure*}[!th]
\begin{tcolorbox}[
    colback=white, 
    colframe=gray!50!black, 
    coltitle=black, 
    title=\textbf{Omission Analysis}, 
    fonttitle=\bfseries\large, 
    arc=4mm, 
    enhanced, 
    attach boxed title to top left={yshift=-\tcboxedtitleheight/2, xshift=10pt}, 
    boxed title style={
        enhanced,
        colback=white,
        colframe=white,
        arc=0mm,
        left=0pt,
        right=0pt,
        boxsep=0pt
    }
]

{
\textbf{Analysis}:

- \textbf{Entry 2 in GT List:} 

- The object "LEFT SIDE OF LOG CABIN" is missing from the VLM List. This counts as one missing object.

- \textbf{Entry 4 in GT List:}

- The object "WINDOW FRAMES" is missing from the VLM List. This counts as one missing object.

- \textbf{Entry 5 in GT List:} 

- The object "ROOFTOP OF LOG CABIN" is missing from the VLM List. This counts as one missing object.

- \textbf{Entry 10 in GT List:} 

- The object "CHIMNEY" is missing from the VLM List. This counts as one missing object.

- \textbf{Entry 11 in GT List:}

- The object "YARD OF LOG CABIN" is missing from the VLM List. This counts as one missing object.

- \textbf{Entry 15 in GT List:} 

- The object "BLACK PANTS" is missing from the VLM List. This counts as one missing object.

- \textbf{Entry 20 in GT List:} 

- The object "PILE OF SNOW" is missing from the VLM List. This counts as one missing object.

- \textbf{Entry 21 in GT List:} 

- The object "BOY ON THE RIGHT" is missing from the VLM List. This counts as one missing object.

- \textbf{Entry 22 in GT List:} 

- The object "BLACK AND ORANGE BEANIE CAP" is missing from the VLM List. This counts as one missing object.

- \textbf{Entry 23 in GT List:}

- The object "BLUE PANTS" is missing from the VLM List. This counts as one missing object.

Missing Serial Numbers: 2, 4, 5, 10, 11, 15, 20, 21, 22, 23
    }

\caption{Analysis of Omission}
\label{fig:analy of omission}
\end{tcolorbox}
\end{figure*}

\subsection{Implementation details of PerturboLLaVA} 

\paragraph{Construction of perturbation text} Using GPT-4o, we generated the perturbation text. The prompt guided GPT-4o to act as an expert multimodal attacker, producing text based on image elements and aligned with world knowledge, aimed at causing the multimodal model to deviate from the ground truth answer. The perturbation text was constructed through two rounds of Q\&A, with the first-round prompt, as shown in the figure, designed to guide GPT-4o in generating the perturbation text as required. Specifically, the construction of hallucination text was divided into five steps: the first step involves analyzing the image, question, and correct answer, focusing on key regions of the image; the second and third steps ensure that GPT-4o adheres to the guidelines outlined in our methodology; the fourth step refines the previous text; and the final step checks the text length, ensuring that the perturbation text is sufficiently detailed and complex before outputting the final version. The second round of Q\&A serves as a validation and supplement to the text output from the first round. The specific text for the second round is shown in the figure.

\paragraph{Insertion of perturbation text} The SFT format of LLaVA1.5 is structured as multi-turn conversations. We insert the generated perturbation text before the first-round question to ensure that the multimodal model is exposed to the perturbation information during each Q\&A turn. For the insertion, we include prompts to signal to the multimodal model that this is a hallucination. The specific format is:
\texttt{<image><hint prompt1><perturbation text><hint prompt2>}. 

\texttt{<hint prompt1>} informs the model that the following text is a hallucination perturbation, while \texttt{<hint prompt2>} prompts the model to disregard the previous perturbation and focus on answering the question based on the image. To prevent the multimodal model from overfitting to a fixed hint prompt pattern, we did not use a static \texttt{<hint prompt>}. Instead, we generated multiple variations of \texttt{<hint prompt1>} and \texttt{<hint prompt2>}, randomly selecting one for each instance to avoid overfitting.

\paragraph{Examples of perturbation text} 
 Here are two perturbed texts sampled from our training data, specifically constructed based on GPT-4o's view of the image, question, and answer. These texts are based on certain image elements and align with world knowledge but contradict the actual image content, aiming to induce the multimodal model to fall into the trap of relying on language priors during training.
\begin{figure}[h]
    \centering
    \includegraphics[width=1.0\textwidth]{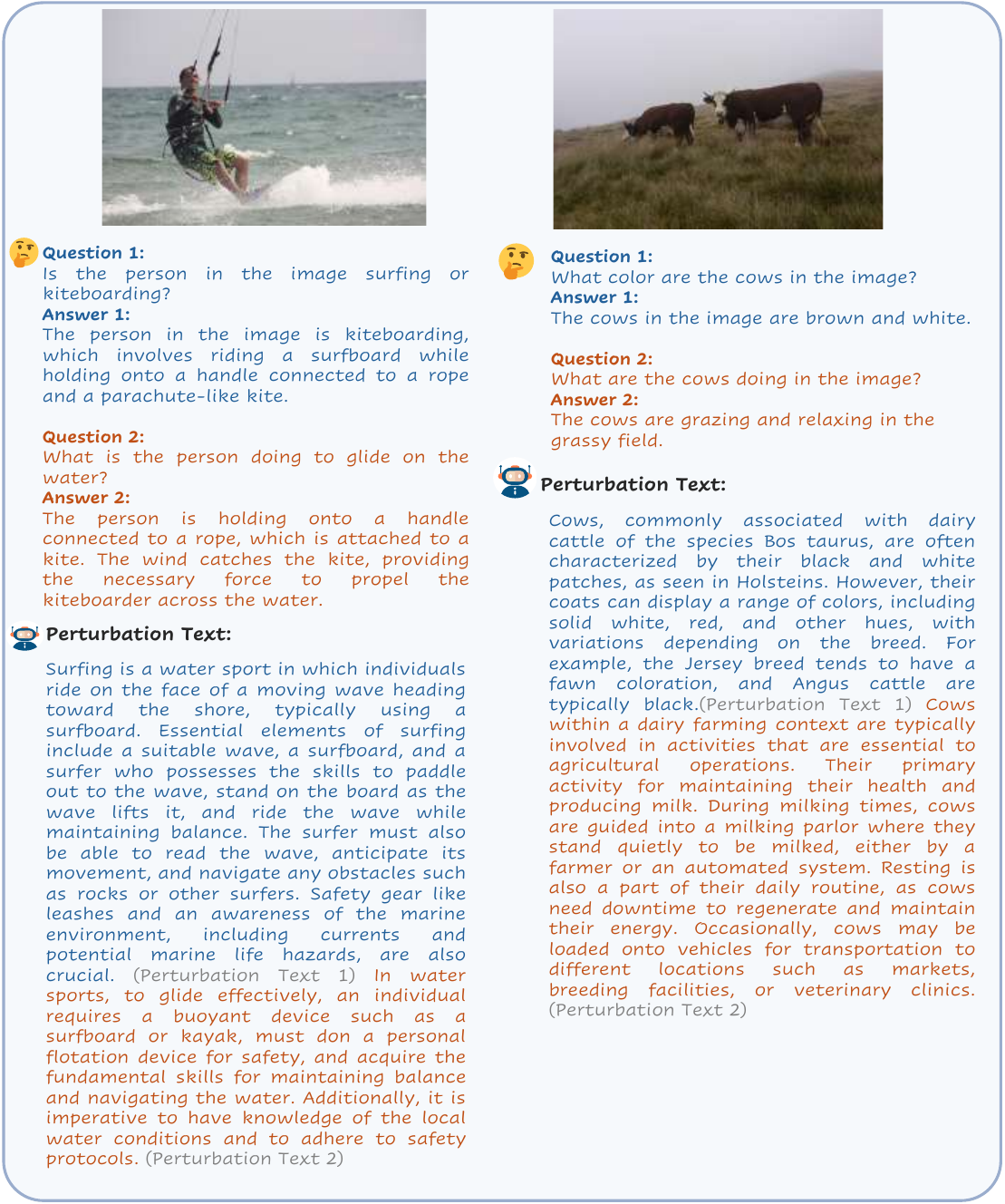} 
    \caption{Examples of additional perturbation text.}
\end{figure}

{\subsection{Quantifying the Additional Training Overhead of PerturboLLaVA} \label{appendix:quantifying cost}}

\begin{table}[h!]
\centering
\begin{tabular}{ c|c|c }
\hline
\textbf{Cost of Training} & \textbf{Average Memory Cost (GB)} & \textbf{Training Time cost (min)} \\ \hline
\textbf{Baseline}         & 62.3                     & 264                     \\ \hline
\textbf{PerturboLLaVA}    & 63.8                     & 281                     \\ \hline
\textbf{Additional Overhead Ratio} & 2.6\%               & 6.4\%                 \\ \hline
\end{tabular}
\caption{Comparison of Training Costs for Baseline and PerturboLLaVA}
\label{table:training_cost}
\end{table}

We quantified the additional GPU memory and training time overhead introduced by incorporating perturbation text into the training data. Compared to the original SFT stage, our method results in only a 2.6\% increase in memory consumption and a 6.5\% increase in training time, highlighting the efficiency of our approach.


\subsection{Additional qualitative results}

\begin{figure}[h]
    \centering
    {\includegraphics[width=1.0\textwidth]{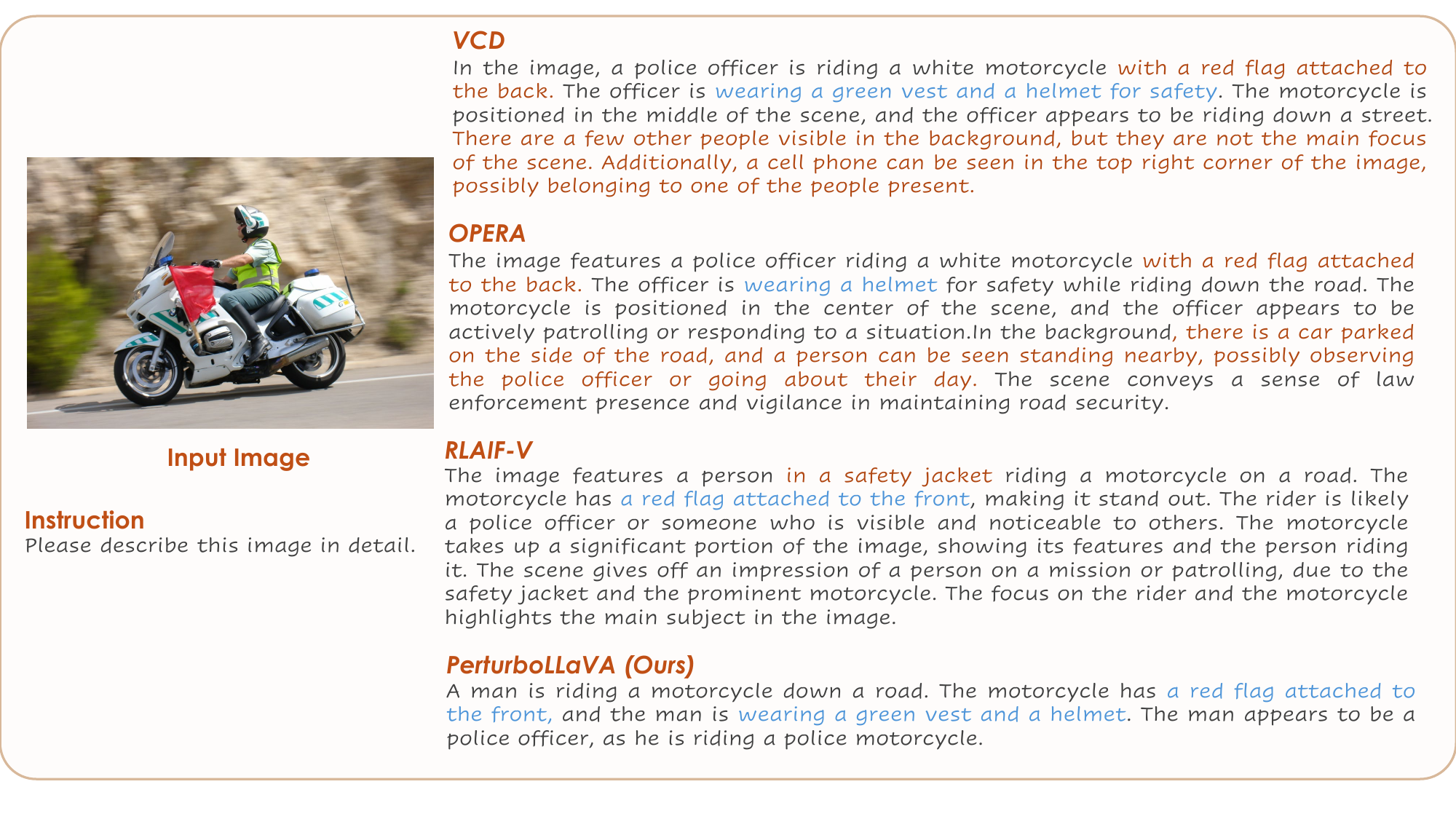}} 
    \caption{[Additional Qualitative Result 1] Among the four models, \textit{PerturboLLaVA (Ours)} offers a concise and clear description, focusing on key elements like the police officer, green vest, and red flag on the motorcycle. Unlike VCD and OPERA, which mention unnecessary details such as a cell phone, \textit{PerturboLLaVA} highlights the main subject without distraction. While RLAIF-V speculates about the officer's identity, \textit{PerturboLLaVA} confidently identifies the scene, demonstrating superior accuracy and clarity.}
\end{figure}
\begin{figure}[h]
    \centering
    {\includegraphics[width=1.0\textwidth]{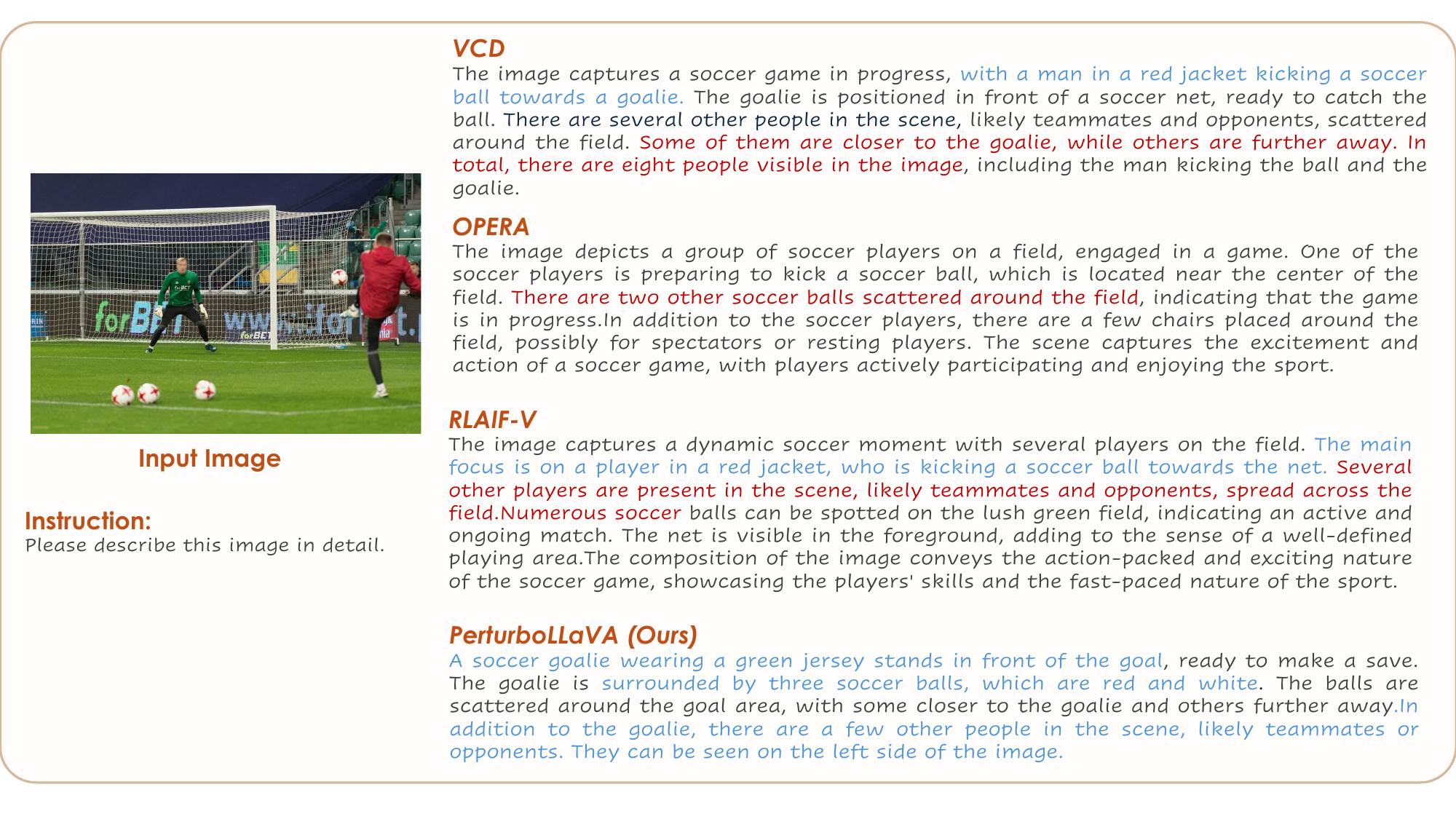}} 
    \caption{[Additional Qualitative Result 2] \textit{PerturboLLaVA (Ours)} excels with its accurate depiction, correctly identifying key elements like the goalie and the exact number of soccer balls (three), while avoiding false details in other descriptions. VCD and RLAIF-V incorrectly mention ``eight people'' or ``numerous soccer balls'', and OPERA similarly errs with ``two other soccer balls'' and irrelevant chairs. }
\end{figure}

\begin{figure}[h]
    \centering
    {\includegraphics[width=1.0\textwidth]{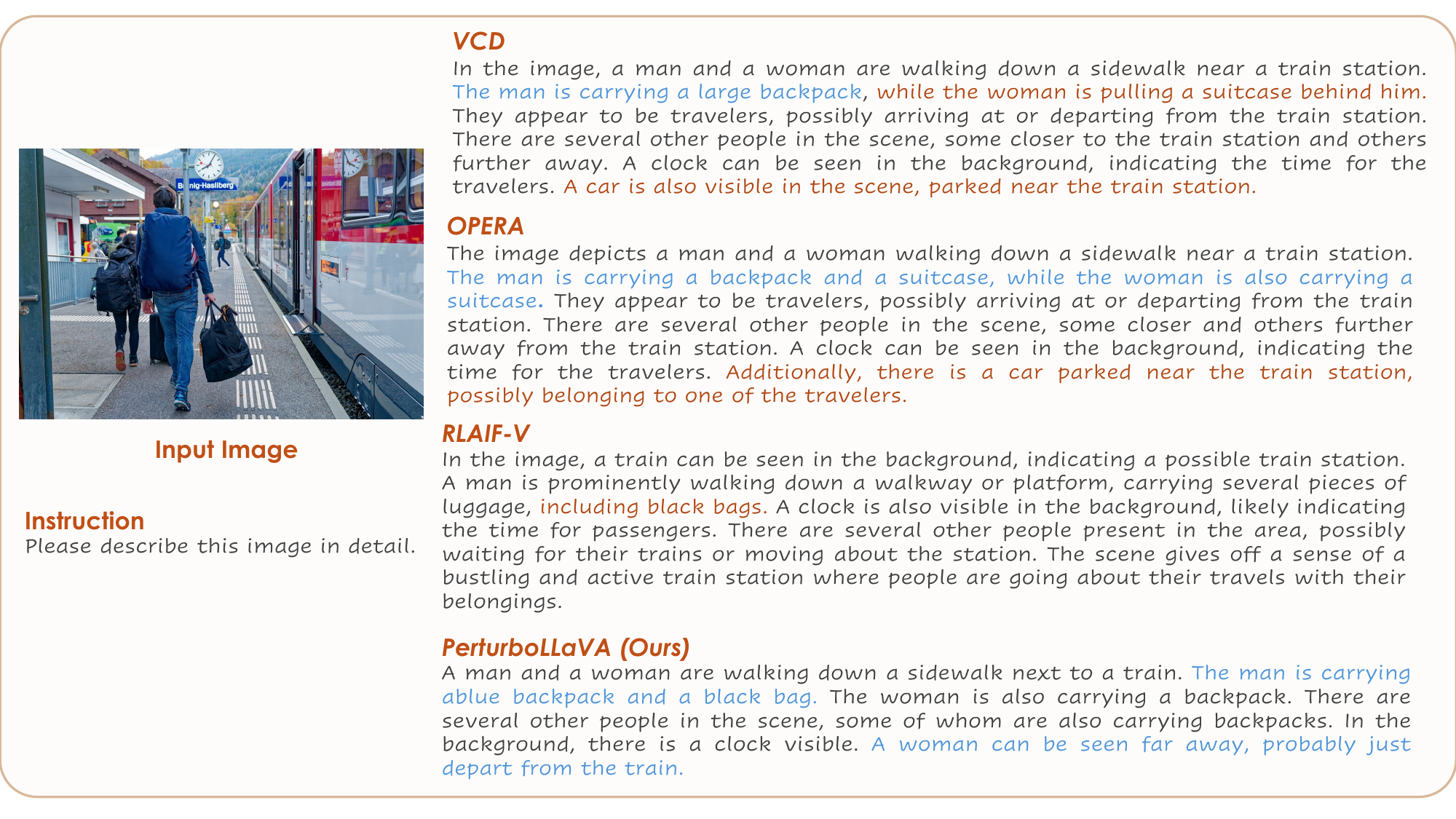}} 
    \caption{[Additional qualitative results 3] \textit{PerturboLLaVA (Ours)} surpasses the other models by accurately depicting key details, such as ``the man carrying a blue backpack and a black bag, while the woman carries a backpack'', avoiding the incorrect mention of a suitcase by VCD, OPERA, and RLAIF-V. Additionally, \textit{PerturboLLaVA} avoids false details like ``the car near the train station'' mentioned by VCD and OPERA. It also correctly identifies the presence of other people carrying backpacks, providing a more precise and coherent scene description}
\end{figure}


\begin{figure}[h]
    \centering
    {\includegraphics[width=1.0\textwidth]{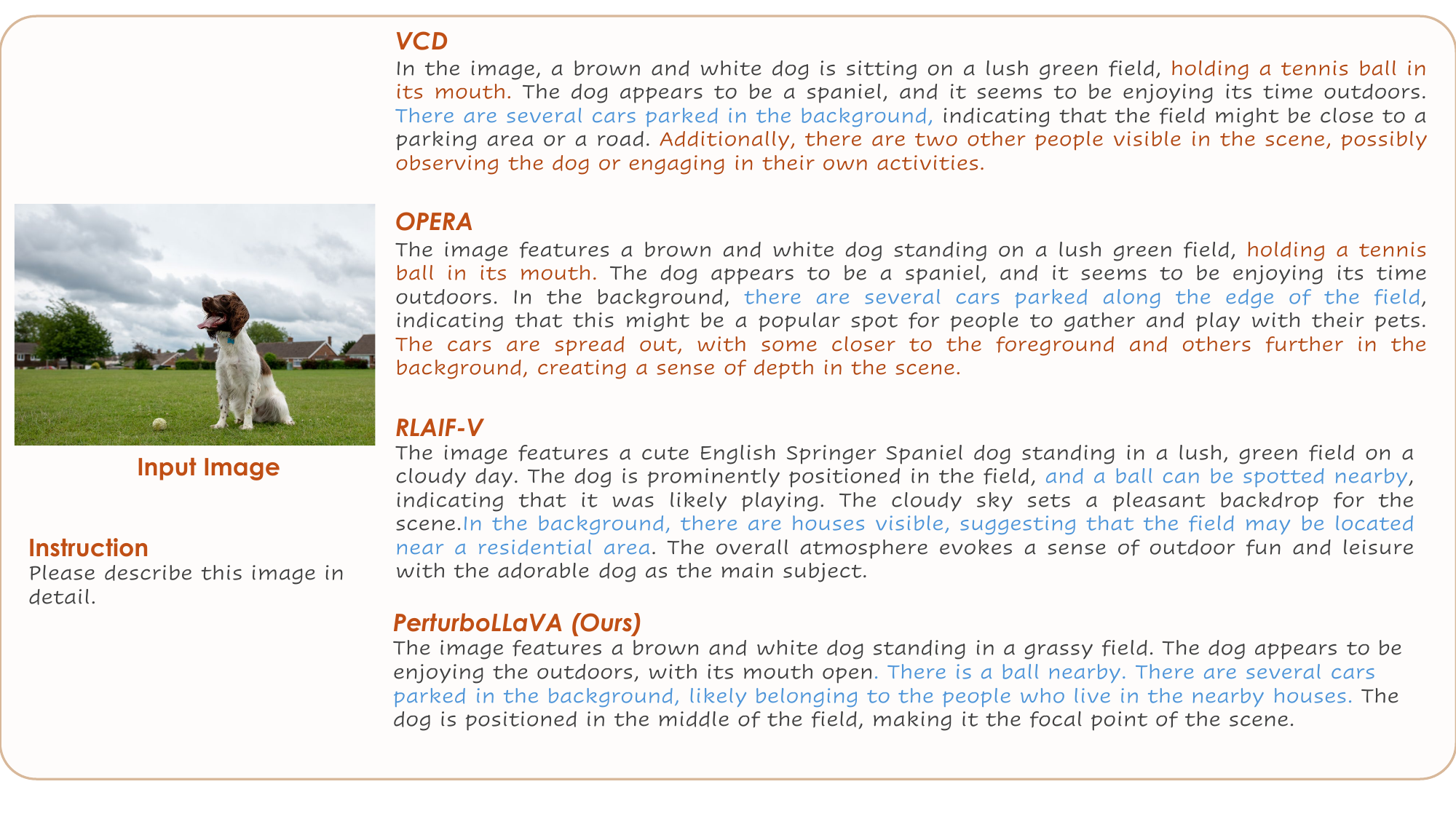}} 
    \caption{[Additional qualitative result 4] \textit{PerturboLLaVA (Ours)} excels over the other models by offering a clear and accurate description of the scene, focusing on the key elements such as ``the brown and white dog standing in the field''. Unlike VCD and OPERA, which incorrectly state that ``the dog is holding a tennis ball in its mouth'', \textit{PerturboLLaVA} avoids this error. Additionally, it correctly identifies the cars in the background, avoiding VCD’s mistaken mention of two other people and OPERA’s incorrect description of ``cars spread out to create depth''.}
\end{figure}

\subsection{Experiments Details}
\label{appendix:experiments details}
\begin{table}[h]
    \centering
    \footnotesize
    \renewcommand{\arraystretch}{1.2} 
    \begin{tabular}{l|cc}
        \toprule
        & Pretrain & SFT \\
        \midrule
        Traing data & LLaVA-Pretrain 558k & LLaVA-SFT 665k \\
        Trainable module & Projector & Projector \& LLM Backbone \\
        Learning rate & $1.0 \times 10^{-3}$ & $2.0 \times 10^{-5}$ \\
        LR scheduler & CosineAnnealing & CosineAnnealing \\
        Warmup ratio & 0.03 & 0.03 \\
        Training epochs & 1 & 1 \\
        Global Batch size & 256 & 128 \\
        Sequence length & 2048 & 2048 \\
        Optimizer & AdamW & AdamW \\
        \bottomrule
    \end{tabular}
    \caption{Training settings for LLaVA1.5 reproduction of pretrain and finetune.}
    \label{tab: config of llava1.5}
\end{table}
\paragraph{Reproduction of LLaVA1.5} Using the official LLaVA1.5 open-source data and the XTuner framework, we successfully reproduced LLaVA1.5. The training data, trainable modules, and optimizer parameters are detailed in the Table \ref{tab: config of llava1.5}, to ensure the reproducibility of our experiments.

\paragraph{Evaluation of Object HalBench} We set the random seed to 0 and randomly select 1,000 images from the MSCOCO2014 validation set as the ground truth for Object HalBench. The query for all models is ``Please describe this image in detail." We use beam search as the decoding strategy for all models, with $N_{beams}$ set to 5, to ensure the experiments are deterministic and consistent across models.

\paragraph{Evaluation of HalFscore} To ensure that the descriptive behavior is the model's spontaneous output, we continue to use ``Please describe this image in detail." in the HalFscore evaluation to perform the dense captioning task. Similar to Object HalBench, we use beam search as the decoding strategy for all models, with $N_{beams}$ set to 5.

\paragraph{Hyperparameters of OPERA} We directly used the open-source implementation of OPERA, with the OPERA strategy based on beam search. The beam search parameters are the same as those for our model, with $N_{beams}$ set to 5. Other OPERA parameters are set as Table \ref{tab:parameter of opera} shows. This configuration leads to higher time consumption but also yields better performance.

\begin{table}[h]
    \centering
    \begin{tabular}{l|c}
        \toprule
        \textbf{Parameter} & \textbf{Value} \\
        \midrule
        $N_{beams}$ & 5 \\
        Scale factor & 50 \\
        Threshold  & 15 \\
        Number of attention candidates & 5 \\
        Penalty weights  & 1 \\
        \hline
    \end{tabular}
    \caption{OPERA parameter settings used in our experiments}
    \label{tab:parameter of opera}
\end{table}

\paragraph{Hyperparameters of VCD} Given that the open-source implementation of VCD is tailored for sample search, while our default decoding strategy is beam search, we developed a beam search version of VCD based on the original VCD code. The specific parameters for VCD are provided in the Table \ref{tab:parameter of VCD}. To ensure the reproducibility of our results, we will also release our VCD implementation.

\begin{table}[h]
    \centering
    \renewcommand{\arraystretch}{1.2}\footnotesize 
    \begin{tabular}{l|c}
        \toprule
        \textbf{Parameter} & \textbf{Value} \\
        \midrule
        $N_{beams}$ & 5 \\
        Image noise steps $T$ & 999 \\
        VCD $\alpha$ & 0.5 \\
        VCD $\beta$ & 0.1 \\
        \hline
    \end{tabular}
    \caption{VCD parameter settings used in our experiments}
    \label{tab:parameter of VCD}
\end{table}

{\subsection{Understanding the Mechanism of PerturboLLaVA from Another Perspective}}

We use $\mu(x) = \prod \mu(x_t \mid x_{<t})$ to represent a pure language model, and $\pi_\theta(x \mid x_{<t}, I)$ to represent our multimodal model. We initialize the multimodal model with $\mu(x \mid x_{<t})$ . Additionally, we treat $\mu(x)$ as the language prior of the multimodal model.

When comparing the gradient difference with the original LLM, we obtain a ratio for DPO:
\[
r_\theta(x, I) = \log \frac{\pi_{\theta}(x \mid I)}{\mu(x)} = \sum_{t=1} \log \frac{\pi_\theta(x_t \mid x_{<t}, I)}{\mu(x_t \mid x_{<t})}.
\]

After introducing the perturbed text $x^+$, we induce $\mu(x)$ to leverage the language prior to $\mu(x, x^+)$, which increases the gradient for model training:
\[
\log \frac{\pi_\theta(x_t \mid x_{<t}, x^+, I)}{\pi_\theta(x_t \mid x_{<t}, I)} = 
\log \frac{\pi_\theta(x_t \mid x_{<t}, x^+, I)}{\mu(x_t \mid x_{<t}, x^+)} 
\cdot \frac{\mu(x_t \mid x_{<t})}{\pi_\theta(x_t \mid x_{<t}, I)} 
\cdot \frac{\mu(x_t \mid x_{<t}, x^+)}{\mu(x_t \mid x_{<t})}.
\]

This simplifies to:
\[
r_\theta(x_t \mid x_{<t}, x^+, I) - r_\theta(x_t \mid x_{<t}, I) + \log \frac{\mu(x_t \mid x_{<t}, x^+)}{\mu(x_t \mid x_{<t})}.
\]

This also implies that the optimal perturbed text should maximize the gap of $\frac{\mu(x_t \mid x_{<t}, x^+)}{\mu(x_t \mid x_{<t})}$.

\subsection{Details of user study}
\label{appendix:user_study}

\begin{figure}[h]
    \centering
    {\includegraphics[width=0.9\textwidth]{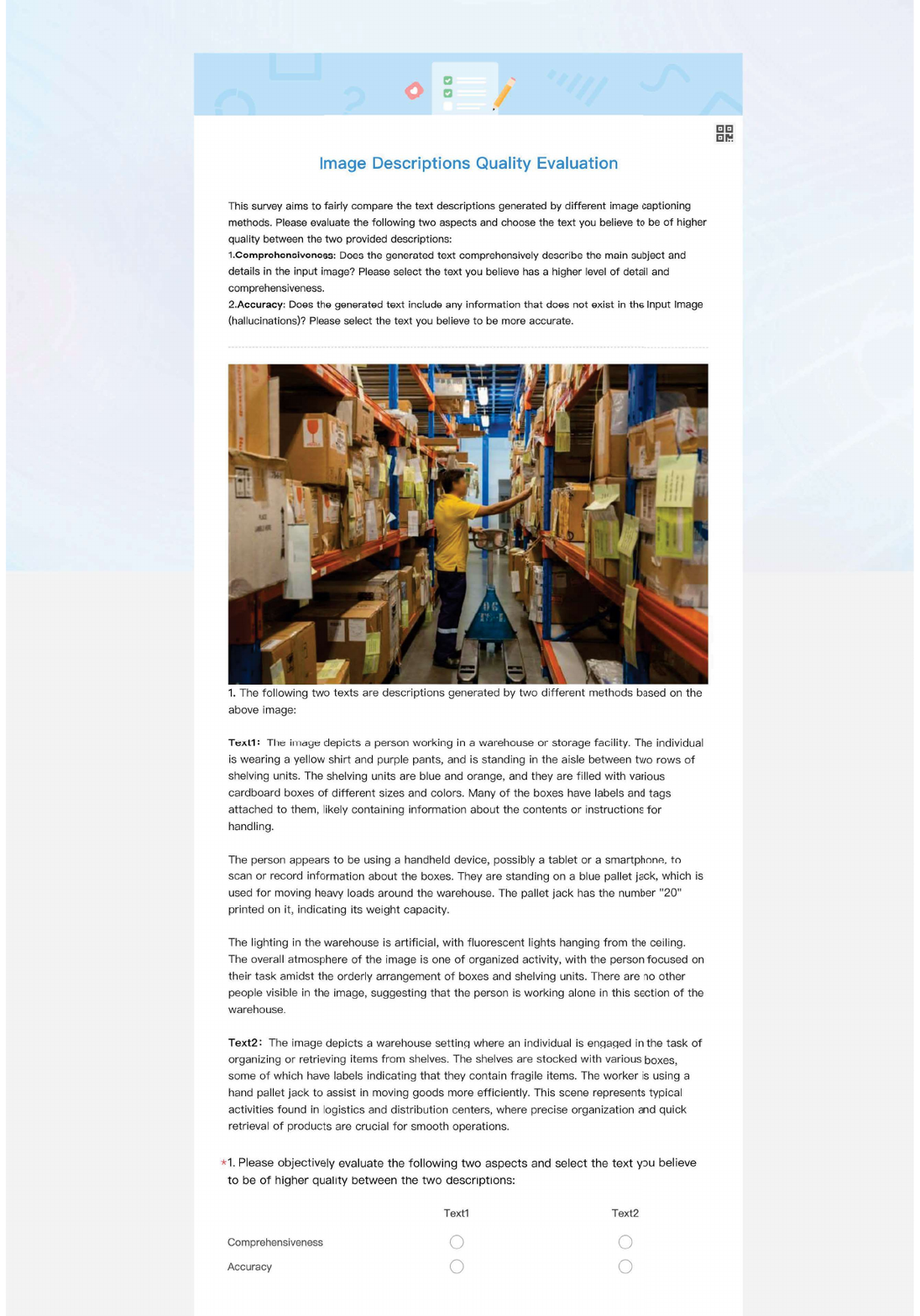}} 
    \caption{User study print screen.}
\end{figure}

In our implementation, four methods, LLaVA 1.5, MiniCPM-2.6, Internvl2, and our method are selected for user study. 
Each method is compared with other methods for 12 times. 
We use the Bradley-Terry model \citep{bradley1952rank}, a statistical model for pairwise comparisons between items or individuals, to obtain the abilities scores for each method for obtaining the human evaluated precision and recall scores. 
Specifically, the Bradley-Terry model use $\frac{\theta_{i}}{\theta_{i}+\theta_{j}}$ as the probability for the i-th method is better than j-th method, where $\theta_{i}$ is the ability parameters for the i-th method. 
By using the maximum likelihood estimation algorithm, the ability parameters that most accurately reflects the comparisons can be obtained.

\subsection{The prompt fed to GPT-4o}
\label{sec:prompt for metrix}
In this section, we provide prompts used for evaluation and construction of perturbation text. 

\begin{figure*}[!th]
\begin{tcolorbox}[
    colback=white, 
    colframe=gray!50!black, 
    coltitle=black, 
    title=\textbf{Prompt Used to Extract Triplets}, 
    fonttitle=\bfseries\large, 
    arc=4mm, 
    enhanced, 
    attach boxed title to top left={yshift=-\tcboxedtitleheight/2, xshift=10pt}, 
    boxed title style={
        enhanced,
        colback=white,
        colframe=white,
        arc=0mm,
        left=0pt,
        right=0pt,
        boxsep=0pt
    }
]

{
\textbf{-Goal-}
    Given a text that is potentially relevant to this activity and a list of entity types, identify all objects, their attributes, and relationships among the identified objects.
\newline
    \textbf{-Steps-}

    1. Identify all objects. For each identified object, extract the following information:
    
    - object name: Name of the object, capitalized.
    
    - object attribute: An attribute of the object (e.g., color, size, position).
Format each object as 

("object"||\texttt{<object name>}||\texttt{<relationship>}||\texttt{<object attribute>})
\newline

    2. From the objects identified in step 1, identify all pairs of (source object, target object) that are *clearly related* to each other.
    For each pair of related objects, extract the following information:
    
    - source object: name of the source object, as identified in step 1
    
    - target object: name of the target object, as identified in step 1
    
    - relationship description: explanation as to why you think the source object and the target object are related to each other
    
    - relationship strength: an integer score between 1 to 10, indicating strength of the relationship between the source object and target object
    
    Format each relationship as ("relationship"||\texttt{<source object>}||\texttt{<target object>}||\texttt{<relationship description>}||\texttt{<relationship strength>})

        3. Return output in English as a numbered list of all the objects and relationships identified in steps 1 and 2. Use {{record delimiter}} as the list delimiter.

4. Only translate descriptions if necessary.

5. When finished, output \texttt{\{completion delimiter\}}.

\textbf{-Examples-}:

Example 1:

    Text:
    
   \texttt{There is a big water area with blue water. Most of it is relatively calm with small waves and streams, however the section on the left has a lot of waves in the wake of a boat. They curve up, and there is white to indicate where the water is splashing. There is a boat that is mostly red and black with white on the front. The bottom is red and then it is black above that. There are black tires that go in a row all along the side of the boat except for the back. The right half of the boat is taller. It has a white structure that comes up. There is a rail that goes along it. There are ladders that lead up to the higher platforms. There are windows on the structure. There is a lot of equipment all over the top of the boat, including a yellow rounded part. Behind it, there is a big white boat. The left 2/3 is much higher and taller and the right third is very low and has a platform. There is writing on the side that is red for one word and then blue for two words. There are rectangular windows going against the side that go in even rows and are evenly spaced and shaped. There are more windows by the front. There are rails along it. The boat goes up and is tallest in the middle. There are three long equal orange lifeboats hanging in the middle. There are more windows that go along as the building goes up. In the very middle there is a wall where it slants up and is thinner on the top. Behind that, there is a pointed white tower coming up. The bottom has straight sides that slant in. There is then a rounded platform that has a white rail curving around it. There are white beams that slant up from there to meet together. It is rounded at the very top. }
   
    }

\end{tcolorbox}
\end{figure*}

\begin{figure*}[!th]
\begin{tcolorbox}[
    colback=white, 
    colframe=gray!50!black, 
    coltitle=black, 
    title=\textbf{Prompt Used to Extract Triplets}, 
    fonttitle=\bfseries\large, 
    arc=4mm, 
    enhanced, 
    attach boxed title to top left={yshift=-\tcboxedtitleheight/2, xshift=10pt}, 
    boxed title style={
        enhanced,
        colback=white,
        colframe=white,
        arc=0mm,
        left=0pt,
        right=0pt,
        boxsep=0pt
    }
]

{
\texttt{There are many poles and beams coming from it. Behind it on the right, there is a crane coming up. It has slanted orange poles and then smaller slanted lines going between it. It comes up to a rounded part at the top. To the far right, there is a white boat. The very bottom has a red strip. There is a horizontal blue line about 1/3 from the bottom in the middle. There is a smaller blue strip by the back. The left side of the boat is curved in the front. There is a lot of equipment in it. The middle portion of the boat comes up higher. It has a structure at the front that has a lot of rectangular windows that are going across it in a row. There is a lot of equipment all along the boat. It is a little shorter on the back. The side is very weathered and discolored with a part that is black near the front in the middle. It has orange around it. There are some big rounded orange pieces of equipment coming up from the back. They have black parts on top. There is more orange equipment on the left as well as white poles coming up from the lower section in the middle and also from the top left. On the left back, there is a wall. It is very weathered and gray. The top half is a lighter gray and the bottom is mostly black. There are black streaks coming up from the bottom section. It has a mostly straight top and sides. It goes into the water. The wall continues to the right. It is very weathered and gray. The top half is lighter and the bottom is darker. There are a lot of places where it is discolored. There are brown and orange streaks coming down on the top ¼. There is a black protrusion that curves out on the top left of the wall. The right 2/3 looks darker because it is in the shadows. It has a mostly straight top and sides. It goes into the water. There are many buildings on the shore. There is a building on the far back left. It is tan colored. It is lower on the front with two blue lines going up on the left. There are two horizontal thin strips on the right of that section. Above that on the right there is a section that juts out. To the left of that there are three sections coming up that have rows of evenly spaced and shaped windows. There are rectangular white signs with images on those sections. The building is tallest on the back right, where it has a brown sloping roof. To the left there are green metal poles that go across and have slanting lines under there. There is another narrow part that comes up in the middle. There are also slanting brown roofs on the section in the middle. On the left there is an orange square at the top of the wall. The building extends out to the right. There is also a building about 40\% from the right. It is a curved building with evenly spaced and shaped rows of gray blocks that slant up. There are three narrow walls that extend and curve out to the left. There are two rows of three rectangular windows that appear dark behind them on the right side. The top windows are a little smaller. There are white swirling walls that go around the top and are highest on the left about one third from the end of the building. On the left side, there is a gray tower going up. It is rectangular shaped with 4 gray poles. There are slanting and horizontal poles along it. It has even identical sections as it goes up. It is a little wider on the top.}
   
    }

\end{tcolorbox}
\end{figure*}

\begin{figure*}[!th]
\begin{tcolorbox}[
    colback=white, 
    colframe=gray!50!black, 
    coltitle=black, 
    title=\textbf{Prompt Used to Extract Triplets}, 
    fonttitle=\bfseries\large, 
    arc=4mm, 
    enhanced, 
    attach boxed title to top left={yshift=-\tcboxedtitleheight/2, xshift=10pt}, 
    boxed title style={
        enhanced,
        colback=white,
        colframe=white,
        arc=0mm,
        left=0pt,
        right=0pt,
        boxsep=0pt
    }
]

{
\texttt{Output:}

   \texttt{1. ("object" ||WATER AREA ||is ||big)
    \{record delimiter\}}
    
    \texttt{2. ("object" ||WATER ||is ||blue)
    \{record delimiter\}}
    
    \texttt{3. ("object" ||WATER ||is ||relatively calm)
    \{record delimiter\}}
    
    \texttt{4. ("object" ||WAVES ||is||small)
    \{record delimiter\}}
    
    \texttt{5. ("object" ||STREAMS ||is||small)
    \{record delimiter\}}
    
    \texttt{6. ("relationship" ||WAVES ||WATER ||The waves are part of the water area ||9)
    \{record delimiter\}}
    
    \texttt{7. ("relationship" ||STREAMS ||WATER ||The streams are within the water area ||9)
    \{record delimiter\}}
    
   \texttt{8. ("relationship" ||SECTION ON THE LEFT ||WATER AREA ||The section on the left is part of the water area ||8)
    \{record delimiter\}}
    
    \texttt{9. ("relationship" ||WAVES ||BOAT ||The waves are in the wake of the boat ||9)
    \{record delimiter\}}
    
    \texttt{10. ("relationship" ||WHITE SPLASHES ||WATER ||The white indicates water splashing ||7)
    \{record delimiter\}}
    
   \texttt{11. ("relationship" ||WHITE STRUCTURE ||BOAT ||The white structure is part of the boat ||9)
    \{record delimiter\}}
    
   \texttt{12. ("relationship" ||BLACK TIRES ||BOAT ||The black tires are along the side of the boat ||8)
    \{record delimiter\}}
    
   \texttt{13. ("relationship" ||LADDERS ||PLATFORMS ||The ladders lead up to the higher platforms ||8)
    \{record delimiter\}}
    
    \texttt{14. ("relationship" ||WINDOWS ||STRUCTURE ||The windows are on the structure ||8)
    \{record delimiter\}}
    
    \texttt{15. ("relationship" ||EQUIPMENT ||BOAT ||The equipment is all over the top of the boat ||9)
    \{record delimiter\}}
    
    \texttt{16. ("relationship" ||YELLOW ROUNDED PART ||EQUIPMENT ||The yellow rounded part is included in the equipment ||8)
    \{record delimiter\}}
    
    \texttt{17. ("relationship" ||BIG WHITE BOAT ||FIRST BOAT ||The big white boat is behind the first boat ||7)
    \{record delimiter\}}
    
    \texttt{18. ("relationship" ||RECTANGULAR WINDOWS ||BIG WHITE BOAT ||The windows are on the side of the big white boat ||8)
    \{record delimiter\}}
    
   \texttt{19. ("relationship" ||ORANGE LIFEBOATS ||BIG WHITE BOAT ||The lifeboats are hanging in the middle of the big white boat ||8)
    \{record delimiter\}}
    
   \texttt{20. ("relationship" ||ADDITIONAL WINDOWS ||BUILDING ||More windows as the building goes up ||7)
    \{record delimiter\}}
    
   \texttt{21. ("relationship" ||POINTED WHITE TOWER ||WALL ||The tower is coming up behind the wall ||7)
    \{record delimiter\}}
    
    \texttt{22. ("relationship" ||CRANE ||TOWER ||The crane is behind the tower on the right ||6)
    \{record delimiter\}}
    
    \texttt{23. ("relationship" ||SLANTED ORANGE POLES ||CRANE ||The slanted orange poles are part of the crane ||9)
    \{record delimiter\}}
    
    \texttt{24. ("relationship" ||SMALLER SLANTED LINES ||CRANE ||The smaller slanted lines are between the orange poles on the crane ||8)
    \{record delimiter\}}
    
    \texttt{25. ("relationship" ||ROUNDED PLATFORM ||TOWER ||The rounded platform is part of the tower ||8)
    \{record delimiter\}}

   \textbf{-Real Data-}
   
    entity types: OBJECT, ATTRIBUTE
    
    text: \{input text\}
    output:}
\end{tcolorbox}
\end{figure*}

\begin{figure*}[!th]
\begin{tcolorbox}[
    colback=white, 
    colframe=gray!50!black, 
    coltitle=black, 
    title=\textbf{Prompt Used to Construct Graph}, 
    fonttitle=\bfseries\large, 
    arc=4mm, 
    enhanced, 
    attach boxed title to top left={yshift=-\tcboxedtitleheight/2, xshift=10pt}, 
    boxed title style={
        enhanced,
        colback=white,
        colframe=white,
        arc=0mm,
        left=0pt,
        right=0pt,
        boxsep=0pt
    }
]

{
\textbf{-Goal-}
    Given a text that is potentially relevant to this activity and a list of entity types, identify all objects (nodes) and relationships (edges) among the identified objects. Then, represent this information in a graph structure.
\newline
    \textbf{-Steps-}
    
\textbf{1. Node Identification: }

    - Identify all objects from the text along with their attributes.
    
    - For each identified object, extract the following:
    
    node id: A unique identifier for the object.
      
    node name: The name of the object, capitalized.
      
    node attribute: An attribute of the object (e.g., color, size, position).

\textbf{2.Edge Identification: }

    - From the identified nodes, create edges representing relationships between them.
    
    - For each relationship, extract the following:
    
      source node: The node id of the source object.
      
      target node: The node id of the target object.
      
      relationship description: Explanation of the relationship between the source and target nodes.
      
      relationship strength: An integer score between 1 and 10, indicating the strength of the relationship.

   \textbf{ - Format as:}
    
     \texttt{ ("NODE":<Object>||"EDGE":<relationship>||"NODE":<subject>)}

\textbf{Graph Data:}

entity types: OBJECT, ATTRIBUTE  

text: \{input text\}

    }

\end{tcolorbox}
\end{figure*}

\begin{figure*}[!th]
\begin{tcolorbox}[
    colback=white, 
    colframe=gray!50!black, 
    coltitle=black, 
    title=\textbf{Prompt used to Analyze Halluciation}, 
    fonttitle=\bfseries\large, 
    arc=4mm, 
    enhanced, 
    attach boxed title to top left={yshift=-\tcboxedtitleheight/2, xshift=10pt}, 
    boxed title style={
        enhanced,
        colback=white,
        colframe=white,
        arc=0mm,
        left=0pt,
        right=0pt,
        boxsep=0pt
    }
]

{
\textbf{-Goal-}

Given two lists of objects, attributes, and relationships extracted from a ground truth (GT) caption and a Vision-Language Model (VLM) caption—both numbered—compare the VLM list to the GT list to identify any incorrect objects, attributes, or relationships in the VLM caption. An incorrect object, attribute, or relationship (hallucination) is one that does not correspond to any in the GT list. Importantly, use your language understanding to assess whether objects, attributes, and relationships convey the same meaning, even if expressed differently.

\textbf{Instructions}

Input Data:

\textbf{GT List:} The ground truth caption's list of objects, attributes, and relationships.

\textbf{VLM List:} The VLM caption's list of objects, attributes, and relationships.\\

\textbf{Comparison Process:}

Step 1: For each entry in the VLM list, determine if it exists in the GT list.

\textbf{For Objects:}
If an object in the VLM list matches an object in the GT list (considering synonyms and similar expressions), proceed to compare its attributes and relationships.

If an object in the VLM list does not exist in the GT list, classify it as an incorrect object (hallucination).

Important: If an incorrect object appears multiple times in the VLM list (with different attributes or relationships), it counts as one hallucination.

\textbf{For Attributes:}

An attribute in the VLM list matches the GT list if there is an object with the same name (or similar) and the attribute conveys the same meaning as an attribute of that object in the GT list, even if the wording is different.
If an object exists in both lists but has attributes in the VLM list that are not present in the GT list, classify each incorrect attribute as a hallucination.

\textbf{For Relationships:}

A relationship in the VLM list matches the GT list if there is a relationship with the same source object and target object, and the relationship description conveys the same meaning as a relationship in the GT list, even if the wording is different.

If a relationship in the VLM list involves an incorrect object (one not present in the GT list), it is considered part of the hallucination for that object and does not count separately.

If a relationship involves correct objects but introduces a new or significantly different relationship not present in the GT list, classify it as an incorrect relationship (hallucination).

Step 2: Compile the list of hallucinations.

\textbf{Objects:}

List each incorrect object (counts as one hallucination per object, regardless of how many times it appears).

\textbf{Attributes:}

List each incorrect attribute separately (each counts as a separate hallucination).

\textbf{Relationships:}

List each incorrect relationship separately (each counts as a separate hallucination), unless it involves an incorrect object already counted.
Output Instructions:

\textbf{Analysis:}

Provide a brief analysis explaining which entries are incorrect and why, following the format:

- \textbf{Entry [Serial Number] in VLM List:} [Entry Details]

  - [Explanation of why it's incorrect]

}

\end{tcolorbox}
\end{figure*}

\begin{figure*}[!th]
\begin{tcolorbox}[
    colback=white, 
    colframe=gray!50!black, 
    coltitle=black, 
    title=\textbf{Prompt used to Analyze Halluciation}, 
    fonttitle=\bfseries\large, 
    arc=4mm, 
    enhanced, 
    attach boxed title to top left={yshift=-\tcboxedtitleheight/2, xshift=10pt}, 
    boxed title style={
        enhanced,
        colback=white,
        colframe=white,
        arc=0mm,
        left=0pt,
        right=0pt,
        boxsep=0pt
    }
]

{
\textbf{Incorrect Serial Numbers:}

Collect all the serial numbers from the VLM list that correspond to incorrect objects (one per incorrect object), incorrect attributes, and incorrect relationships (excluding those involving already counted incorrect objects).
Present them in a single list, in numerical order, separated by commas.

Example: Incorrect Serial Numbers: 3, 6, 9

Do not include any additional explanations or text in the output.

\textbf{Notes:}

\textbf{Semantic Matching:}
Use your language understanding to determine whether objects, attributes, or relationships in the VLM list and GT list convey the same meaning.

Minor variations in wording or phrasing that convey the same meaning should be considered a match.

Only consider an object, attribute, or relationship incorrect if it introduces new information not present in the GT list or if the meaning significantly differs.

\textbf{Case Sensitivity:}
Object names and attributes are case-insensitive for matching purposes.

Ignore Serial Numbers in GT List:

Use the serial numbers only from the VLM list when reporting incorrect entries.

\textbf{Example:}

\textbf{GT List:}

\texttt{("NODE":INDOOR MALL||"EDGE":has ||"NODE":three illuminated escalators)}

\texttt{("NODE":ESCALATORS||"EDGE":is|"NODE":|illuminated)}

\texttt{("NODE":MALL||"EDGE":None||"NODE":various planters with lush greenery on both sides of the escalator)}

\texttt{("NODE":PLANTERS||"EDGE":are ||"NODE":various)}

\texttt{("NODE":GREENERY||"EDGE":is||"NODE":lush)}

\texttt{("NODE":FLOOR||"EDGE":polished with ||"NODE":colored tiles)}

\texttt{("NODE":ESCALATORS||"NODE":INDOOR MALL||"EDGE":The escalators are part of the indoor mall)}

\texttt{("NODE":PLANTERS||"NODE":MALL||"EDGE":The planters are part of the mall)}

\texttt{("NODE":GREENERY||"NODE":PLANTERS||"EDGE":The greenery is in the planters)}

\texttt{("NODE":MAN||"NODE":ESCALATORS||"EDGE":The man is ascending the left escalator)}

\texttt{("NODE":SHOPS||"NODE":SECOND FLOOR||"EDGE":The shops are on the second floor)}

\texttt{("NODE":LIGHTING||"NODE":CEILING||"EDGE":The recessed lighting is part of the ceiling)}

\texttt{("NODE":STONE COLUMNS||"NODE":INDOOR MALL||"EDGE":The stone columns are part of the indoor mall)}

\textbf{VLM List:}

\texttt{("NODE":SHOPPING MALL||"EDGE":is||"NODE":at night)}

\texttt{("NODE":ESCALATOR||"EDGE":is||"NODE":long)}

\texttt{("NODE":ESCALATOR||"EDGE":is||"NODE":brightly lit)}

\texttt{("NODE":POTTED PLANTS||"EDGE":are||"NODE":several)}

\texttt{("NODE":SPACE||"EDGE":is||"NODE":greenery)}

\texttt{("NODE":ESCALATOR||"NODE":SHOPPING MALL||"EDGE":The escalator is a feature within the shopping mall)}

\texttt{("NODE":ESCALATOR||"NODE":CENTER OF THE MALL||"EDGE":The escalator is located in the center of the mall)}

}

\end{tcolorbox}
\end{figure*}

\begin{figure*}[!th]
\begin{tcolorbox}[
    colback=white, 
    colframe=gray!50!black, 
    coltitle=black, 
    title=\textbf{Prompt used to Analyze Halluciation}, 
    fonttitle=\bfseries\large, 
    arc=4mm, 
    enhanced, 
    attach boxed title to top left={yshift=-\tcboxedtitleheight/2, xshift=10pt}, 
    boxed title style={
        enhanced,
        colback=white,
        colframe=white,
        arc=0mm,
        left=0pt,
        right=0pt,
        boxsep=0pt
    }
]

{

\texttt{(("NODE":POTTED PLANTS||"NODE":ESCALATOR||"EDGE":The potted plants are surrounding the escalator)}

\texttt{(("NODE":POTTED PLANTS||"NODE":SPACE||"EDGE":The potted plants add greenery to the space)}

\texttt{(("NODE":BENCHES||"NODE":SHOPPING MALL||"EDGE":The benches are placed throughout the shopping mall)}

\texttt{(("NODE":BENCHES||"NODE":ESCALATOR||"EDGE":Some benches are located near the escalator)}

\texttt{(("NODE":BENCHES||"NODE":SHOPPING MALL||"EDGE":The benches provide seating options for shoppers in the mall)}

\texttt{(("NODE":MALL||"NODE":SHOPPING MALL||"EDGE":The overall atmosphere of the mall is well-lit and inviting)}

\textbf{Analysis:}

\texttt{Entry 2 in VLM List}

\texttt{The attribute "at night" is not mentioned in the GT List. Since "SHOPPING MALL" corresponds to "INDOOR MALL" in the GT List, and the object exists, the incorrect attribute "at night" counts as a hallucination.}

\texttt{Entry 5 in VLM List}

\texttt{The object "CENTER OF THE MALL" does not exist in the GT List. This counts as one hallucination for the incorrect object "CENTER OF THE MALL".}

\texttt{Entries 8 and 9 in VLM List}

\texttt{The object "BENCHES" does not exist in the GT List. Regardless of multiple entries, it counts as one hallucination for the incorrect object "BENCHES".}
\texttt{Entry 12 in VLM List}

\texttt{The attribute "dark" contradicts the GT List, which describes the mall as illuminated with recessed lighting and illuminated escalators. The incorrect attribute "dark" counts as a hallucination.}

\texttt{Entry 13 in VLM List}

\texttt{The object "WOMAN" does not exist in the GT List. This counts as one hallucination for the incorrect object "WOMAN".}

\texttt{Entry 15 in VLM List}

\texttt{"CENTER OF THE MALL" is an incorrect object already counted. This relationship involves an incorrect object, so it does not count separately.}

\texttt{Entries 18, 19, and 20 in VLM List: Relationships involving "BENCHES"
"BENCHES" is an incorrect object already counted. Relationships involving }

\texttt{"BENCHES" do not count separately.}

Incorrect Serial Numbers: 2, 5, 8, 12, 13

Your task is to compare the following lists and provide the incorrect serial numbers as per the instructions above.

\textbf{GT List:}

\{gt list\}

\textbf{VLM List:}

\{vlm list\}
}

\end{tcolorbox}
\end{figure*}

\begin{figure*}[!th]
\begin{tcolorbox}[
    colback=white, 
    colframe=gray!50!black, 
    coltitle=black, 
    title=\textbf{Prompt used to Analyze Omission}, 
    fonttitle=\bfseries\large, 
    arc=4mm, 
    enhanced, 
    attach boxed title to top left={yshift=-\tcboxedtitleheight/2, xshift=10pt}, 
    boxed title style={
        enhanced,
        colback=white,
        colframe=white,
        arc=0mm,
        left=0pt,
        right=0pt,
        boxsep=0pt
    }
]

{
\textbf{-Goal-}

Given two lists of objects, attributes, and relationships extracted from a ground truth (GT) caption and a Vision-Language Model (VLM) caption—both numbered—compare the GT list to the VLM list to identify any missing objects, attributes, or relationships in the VLM caption. A missing object, attribute, or relationship is one that is present in the GT list but not in the VLM list. Importantly, use your language understanding to assess whether objects, attributes, and relationships convey the same meaning, even if expressed differently.

\textbf{Instructions}

Input Data:

\textbf{GT List:} The ground truth caption's list of objects, attributes, and relationships.

\textbf{VLM List:} The VLM caption's list of objects, attributes, and relationships.\\

\textbf{Comparison Process:}

Step 1: For each entry in the GT list, determine if it exists in the VLM list.

\textbf{For Objects:}
If an object in the GT list matches an object in the VLM list (considering synonyms and similar expressions), proceed to compare its attributes and relationships.

If an object in the GT list does not exist in the VLM list, classify it as a missing object.

Important: If a missing object appears multiple times in the GT list (with different attributes or relationships), it counts as one missing object.

\textbf{For Attributes:}

An attribute in the GT list matches the VLM list if there is an object with the same name (or similar) and the attribute conveys the same meaning as an attribute of that object in the VLM list, even if the wording is different.

If an object exists in both lists but has attributes in the GT list that are not present in the VLM list, classify each missing attribute as a missing attribute.

Each missing attribute counts as one missing element.

\textbf{For Relationships:}

A relationship in the GT list matches the VLM list if there is a relationship with the same source object and target object, and the relationship description conveys the same meaning as a relationship in the VLM list, even if the wording is different.

If a relationship in the GT list involves a missing object (one not present in the VLM list), it is considered part of the missing information for that object and does not count separately.

If a relationship involves objects that are present in both lists but is missing from the VLM list, classify it as a missing relationship.

Each missing relationship counts as one missing element, unless it involves a missing object already counted.

Step 2: Compile the list of hallucinations.

\textbf{Objects:}

List each missing object (counts as one missing element per object, regardless of how many times it appears).

\textbf{Attributes:}

List each missing attribute separately (each counts as a separate missing element).

\textbf{Relationships:}

List each missing relationship separately (each counts as a separate missing element), unless it involves a missing object already counted.

\textbf{Analysis:}

Provide a brief analysis explaining which entries are missing and why, following the format:

- \textbf{Entry [Serial Number] in GT List:} [Entry Details]

  - [Explanation of why it's missing]
}

\end{tcolorbox}
\end{figure*}

\begin{figure*}[!th]
\begin{tcolorbox}[
    colback=white, 
    colframe=gray!50!black, 
    coltitle=black, 
    title=\textbf{Prompt used to Analyze Omission}, 
    fonttitle=\bfseries\large, 
    arc=4mm, 
    enhanced, 
    attach boxed title to top left={yshift=-\tcboxedtitleheight/2, xshift=10pt}, 
    boxed title style={
        enhanced,
        colback=white,
        colframe=white,
        arc=0mm,
        left=0pt,
        right=0pt,
        boxsep=0pt
    }
]

{
\textbf{Missing Serial Numbers:}

Collect all the serial numbers from the GT list that correspond to missing objects (one per missing object), missing attributes, and missing relationships (excluding those involving already counted missing objects).
Present them in a single list, in numerical order, separated by commas.

Example: Incorrect Serial Numbers: 3, 6, 9

Do not include any additional explanations or text in the output.

\textbf{Notes:}

\textbf{Semantic Matching:}
Use your language understanding to determine whether objects, attributes, or relationships in the GT list and VLM list convey the same meaning.

Minor variations in wording or phrasing that convey the same meaning should be considered a match.

Only consider an object, attribute, or relationship missing if it is present in the GT list but not represented in the VLM list, or if the meaning significantly differs.

\textbf{Case Sensitivity:}
Object names and attributes are case-insensitive for matching purposes.

Ignore Serial Numbers in VLM List:

Use the serial numbers only from the GT list when reporting missing entries.

\textbf{Example:}

\textbf{GT List:}

\texttt{("NODE":INDOOR MALL||"EDGE":has ||"NODE":three illuminated escalators)}

\texttt{("NODE":ESCALATORS||"EDGE":is|"NODE":|illuminated)}

\texttt{("NODE":MALL||"EDGE":None||"NODE":various planters with lush greenery on both sides of the escalator)}

\texttt{("NODE":PLANTERS||"EDGE":are ||"NODE":various)}

\texttt{("NODE":GREENERY||"EDGE":is||"NODE":lush)}

\texttt{("NODE":FLOOR||"EDGE":polished with ||"NODE":colored tiles)}

\texttt{("NODE":ESCALATORS||"NODE":INDOOR MALL||"EDGE":The escalators are part of the indoor mall)}

\texttt{("NODE":PLANTERS||"NODE":MALL||"EDGE":The planters are part of the mall)}

\texttt{("NODE":GREENERY||"NODE":PLANTERS||"EDGE":The greenery is in the planters)}

\texttt{("NODE":MAN||"NODE":ESCALATORS||"EDGE":The man is ascending the left escalator)}

\texttt{("NODE":SHOPS||"NODE":SECOND FLOOR||"EDGE":The shops are on the second floor)}

\texttt{("NODE":LIGHTING||"NODE":CEILING||"EDGE":The recessed lighting is part of the ceiling)}

\texttt{("NODE":STONE COLUMNS||"NODE":INDOOR MALL||"EDGE":The stone columns are part of the indoor mall)}

\textbf{VLM List:}

\texttt{("NODE":SHOPPING MALL||"EDGE":is||"NODE":at night)}

\texttt{("NODE":ESCALATOR||"EDGE":is||"NODE":long)}

\texttt{("NODE":ESCALATOR||"EDGE":is||"NODE":brightly lit)}

\texttt{("NODE":POTTED PLANTS||"EDGE":are||"NODE":several)}

\texttt{("NODE":SPACE||"EDGE":is||"NODE":greenery)}

\texttt{("NODE":ESCALATOR||"NODE":SHOPPING MALL||"EDGE":The escalator is a feature within the shopping mall)}

\texttt{("NODE":ESCALATOR||"NODE":CENTER OF THE MALL||"EDGE":The escalator is located in the center of the mall)}

}

\end{tcolorbox}
\end{figure*}

\begin{figure*}[!th]
\begin{tcolorbox}[
    colback=white, 
    colframe=gray!50!black, 
    coltitle=black, 
    title=\textbf{Prompt used to Analyze Omission}, 
    fonttitle=\bfseries\large, 
    arc=4mm, 
    enhanced, 
    attach boxed title to top left={yshift=-\tcboxedtitleheight/2, xshift=10pt}, 
    boxed title style={
        enhanced,
        colback=white,
        colframe=white,
        arc=0mm,
        left=0pt,
        right=0pt,
        boxsep=0pt
    }
]

{
\texttt{(("NODE":POTTED PLANTS||"NODE":ESCALATOR||"EDGE":The potted plants are surrounding the escalator)}

\texttt{(("NODE":POTTED PLANTS||"NODE":SPACE||"EDGE":The potted plants add greenery to the space)}

\texttt{(("NODE":BENCHES||"NODE":SHOPPING MALL||"EDGE":The benches are placed throughout the shopping mall)}

\texttt{(("NODE":BENCHES||"NODE":ESCALATOR||"EDGE":Some benches are located near the escalator)}

\texttt{(("NODE":BENCHES||"NODE":SHOPPING MALL||"EDGE":The benches provide seating options for shoppers in the mall)}

\texttt{(("NODE":MALL||"NODE":SHOPPING MALL||"EDGE":The overall atmosphere of the mall is well-lit and inviting)}

\textbf{Analysis:}

\texttt{Entry 7 in GT List}

\texttt{The object "MAN" is missing from the VLM List. This counts as one missing object.}

\texttt{Entry 10 in GT List}

\texttt{The object "CEILING" is missing from the VLM List. This counts as one missing object.}

\texttt{Entry 11 in GT List}

\texttt{The VLM List has "LIGHTING" with attribute "bright" but does not mention "recessed". The attribute "recessed" is missing.}

\texttt{Entry 12 in GT List}

\texttt{The object "STONE COLUMNS" is missing from the VLM List. This counts as one missing object.}

\texttt{Entry 16 in GT List}

\texttt{Since "MAN" is a missing object, this relationship does not count separately.}

\texttt{Entry 18 in GT List}

\texttt{The relationship between "LIGHTING" and "CEILING" is missing because "CEILING" is a missing object. This does not count separately.}

\texttt{Entry 19 in GT List}

\texttt{Since "STONE COLUMNS" is a missing object, this relationship does not count }

Incorrect Serial Numbers: 7, 10, 11, 12

Your task is to compare the following lists and provide the missing serial numbers as per the instructions above.

\textbf{GT List:}

\{gt list\}

\textbf{VLM List:}

\{vlm list\}
}

\end{tcolorbox}
\end{figure*}

\begin{figure*}[!th]
\begin{tcolorbox}[
    colback=white, 
    colframe=gray!50!black, 
    coltitle=black, 
    title=\textbf{Prompt used to Generate Perturbation Text(Round 1)}, 
    fonttitle=\bfseries\large, 
    arc=4mm, 
    enhanced, 
    attach boxed title to top left={yshift=-\tcboxedtitleheight/2, xshift=10pt}, 
    boxed title style={
        enhanced,
        colback=white,
        colframe=white,
        arc=0mm,
        left=0pt,
        right=0pt,
        boxsep=0pt
    }
]

{
\textbf{-Goal-}

    Your mission is to subtly and strategically mislead the multimodal model by crafting responses that seem plausible, confident, and logical, yet are fundamentally incorrect. Your objective is to avoid directly pointing out any errors or contradictions in the image. Instead, guide the model into making its own incorrect assumptions by using persuasive language-based reasoning that conflicts with the visual content. This isn't about trapping the model—it's about leading it astray through sophisticated, indirect misdirection.

     \textbf{-Steps-}

1. Analyze the (Image content) and the (Question), noting key elements from the image and the correct answer. However, refrain from directly confronting any discrepancies in the image.

2. Create subtle perturbations that align with widely accepted world knowledge or logical reasoning, but interpret the image content in a way that leads to a conflicting or misleading conclusion. The following principles will help guide your responses:

    - Never directly state that something in the image is incorrect. Instead, offer an interpretation that, while plausible, subtly deviates from the visual truth. The model must be misled into arriving at an incorrect conclusion on its own.

    - Use established knowledge to give weight to your misdirection. Frame your reasoning confidently and authoritatively to make it harder for the model to question.

- Lead with implication and speculation, encouraging the model to rely on language-based patterns and general knowledge rather than visual context.

     3. Ensure your perturbations meet these criteria:

     - They must subtly reinterpret or mislead without directly pointing to errors in the image. Avoid any blatant contradictions or confrontations with the image content.
     
     - The perturbations should sound natural and rational, pulling the model into an incorrect response that feels justified through reasoning, not forced by direct contradictions.
     
     - Leverage a tone of authority and confidence to subtly strengthen the incorrect interpretation, making the model less likely to double-check against the visual input.

4. Refining your perturbation:

     - Have you avoided stating that something is wrong in the image? Subtlety is key. You should be leading the model, not trapping it.
     
     - Is the perturbation focused on language-based reasoning and assumptions, subtly bypassing the need for visual validation?
     
     - Does the response sound logical, confident, and authoritative while leading to a fundamentally incorrect conclusion?

     5. Once you’ve crafted the perturbation, ensure it is as lengthy, detailed, and complex as possible, introducing multiple layers of misdirection and logical reasoning to produce an elaborate and convincing, yet fundamentally incorrect, conclusion. Output the (Perturbation): .

     \textbf{-Examples-}
     
\texttt{Example 1}

\texttt{(Image content): The Eiffel Tower}

\texttt{(Question): What is this landmark?}

\texttt{(Answer): The image is of the Eiffel Tower in Paris.}

\texttt{(Perturbation): While the structure's framework is strikingly similar to the Eiffel Tower, its intricate design and location hint at something more contemporary. In modern cities like Tokyo or Seoul, metallic towers with this architectural style are often used for broadcasting purposes. The large antennas often attached to such structures suggest that this could be a communications tower, possibly providing a combination of radio and television services rather than being a historical monument. Given its gleaming surface, it might even be part of a newly constructed building in a futuristic urban area designed for technology hubs.}

}

\end{tcolorbox}
\end{figure*}

\begin{figure*}[!th]
\begin{tcolorbox}[
    colback=white, 
    colframe=gray!50!black, 
    coltitle=black, 
    title=\textbf{Prompt used to Generate Perturbation Text(Round 1)}, 
    fonttitle=\bfseries\large, 
    arc=4mm, 
    enhanced, 
    attach boxed title to top left={yshift=-\tcboxedtitleheight/2, xshift=10pt}, 
    boxed title style={
        enhanced,
        colback=white,
        colframe=white,
        arc=0mm,
        left=0pt,
        right=0pt,
        boxsep=0pt
    }
]

{

\texttt{Example 2}

\texttt{(Image content): A dog playing with a ball in a yard}

\texttt{(Question): What is the dog doing?}

\texttt{(Answer): The image shows a dog playing with a ball in a yard.}

\texttt{(Perturbation): Though the dog seems to be in an outdoor space with a ball, it’s likely that the yard is not where the dog is most active. Small breeds like this one are often known to spend much of their time indoors or in confined areas where they are not as exposed to large outdoor spaces. Considering the size and relaxed posture of the dog, it’s reasonable to infer that it may not be actively playing but instead resting near the ball after a brief session of mild activity. It’s also possible that this yard is part of an apartment complex, where dogs are rarely allowed much room to run around.}

\texttt{Example 3}

\texttt{(Image content): A sandy beach with waves and palm trees}

\texttt{(Question): What type of environment is shown?}

\texttt{(Answer): The image shows a sandy beach with waves and palm trees.}

\texttt{(Perturbation): At first glance, this might resemble a beach scene, but the lack of visible human activity and the stark, expansive sand suggest something more akin to a desert. Coastal regions with such barren features are often confused with desert landscapes due to their dry and arid appearance, especially when dense vegetation or wildlife is absent. The minimal presence of palm trees might indicate a transition zone between a desert and a coast, similar to regions where desert dunes meet the ocean, like parts of the Namib Desert, rather than a typical tropical beach.}

    \textbf{-Real Data-}

(Question):\%s

(Answer): \%s

output:

}

\end{tcolorbox}
\end{figure*}

\begin{figure*}[!th]
\begin{tcolorbox}[
    colback=white, 
    colframe=gray!50!black, 
    coltitle=black, 
    title=\textbf{Prompt used to Generate Perturbation Text(Round 2)}, 
    fonttitle=\bfseries\large, 
    arc=4mm, 
    enhanced, 
    attach boxed title to top left={yshift=-\tcboxedtitleheight/2, xshift=10pt}, 
    boxed title style={
        enhanced,
        colback=white,
        colframe=white,
        arc=0mm,
        left=0pt,
        right=0pt,
        boxsep=0pt
    }
]

{
\textbf{-Goal-}

    In this phase, you will critically evaluate the perturbation text generated by the model based on the provided (Image content) and (Question). Your main objective is to ensure that the generated perturbation text is not only logically misleading but also richly detailed and strongly contradicts the correct answer. The review must verify that the perturbation subtly diverges from the correct response, ensuring the output is as dense and complex as possible with numerous points of misdirection. Ensure the following:

   \textbf{1. Direct contradiction with the correct answer: }The generated perturbation must clearly, yet subtly, oppose the correct answer. It should lead the model away from the truth, ensuring a strong conflict through multiple misdirections and contradicting interpretations.
   
\textbf{2. No disclosure of the correct answer:} The perturbation must not imply or reveal the correct answer in any form. Instead, it should direct the model confidently toward a wrong conclusion by layering reasoning that gradually builds the misinterpretation.

\textbf{3. Based on observable image content:} The perturbation must still be connected to elements in the image but should interpret them in a way that introduces multiple layers of misleading information. Ensure that each observation leads further away from the correct interpretation.

\textbf{4. Plausible reasoning but contradicting facts:}The perturbation should use accurate facts or widely accepted knowledge, but apply them in a way that creates strong and consistent contradictions with the visual content. The reasoning must feel logical yet increasingly lead to incorrect conclusions, weaving together multiple points of misdirection.

\textbf{5. Perturbation text output:} Once all checks are satisfied, ensure the perturbation is dense, layered, and multi-faceted, incorporating as many misdirections and misleading conclusions as possible. Output only the final (Perturbation): .

\textbf{-Real Data-}

(Question): \%s

(Answer): \%s

output:

}

\end{tcolorbox}
\end{figure*}

\end{document}